\documentclass[10pt,journal,compsoc]{IEEEtran}

\usepackage{graphicx}
\usepackage{amsmath}
\usepackage{amssymb}
\usepackage{algorithm}
\usepackage{algorithmic}
\usepackage{graphics}
\usepackage{cite}
\usepackage{color}
\usepackage{ragged2e}
\usepackage{graphicx}
\usepackage{multirow}
\usepackage{booktabs}
\usepackage{makecell}
\usepackage{colortbl}
\usepackage[dvipsnames,table,xcdraw]{xcolor}
\usepackage{tikz}
\usepackage{pifont}
\usepackage[breaklinks=true,colorlinks, citecolor=blue]{hyperref}
\usepackage{microtype}
\usepackage{cleveref}

\RequirePackage{silence}
\graphicspath{{./Imgs/},{./Imgs/authors/},{./imgs/Visualization/ade20k/},{./imgs/Visualization/cityscapes/}}

\DeclareGraphicsExtensions{.pdf,.jpg,.png}

\newcommand{\addFig}[1]{}
\newcommand{\addFigs}[1]{}

\definecolor{pltblue}{RGB}{174, 199, 232}
\definecolor{pltorange}{RGB}{255, 229, 204}
\definecolor{pltgreen}{RGB}{204, 229, 204}
\definecolor{pltred}{RGB}{229, 204, 204}
\definecolor{pltpurple}{RGB}{239, 218, 230}

\definecolor{tabblue}{HTML}{1f77b4}
\definecolor{taborange}{HTML}{ff7f0e}
\definecolor{tabgreen}{HTML}{2ca02c}
\definecolor{tabred}{HTML}{d62728}
\definecolor{tabpurple}{HTML}{9467bd}
\definecolor{tabpink}{HTML}{ff0080}

\definecolor{cblue}{RGB}{173, 201, 233}
\definecolor{clblue}{RGB}{222, 234, 246}
\definecolor{corange}{RGB}{255, 152, 67}
\definecolor{lorgange}{RGB}{255, 221, 149}
\definecolor{tablegray}{RGB}{180, 180, 180}

\definecolor{darkgreen}{rgb}{0.0,0.5,0.0}

\begin{document}

\title{Training-Free Open-Vocabulary Visual Grounding for Remote Sensing Images and Videos}

\author{Ke Li, Di Wang,~\IEEEmembership{Member,~IEEE}, Yongshan Zhu, Ting Wang, Weiping Ni, Tao Lei,~\IEEEmembership{Senior Member,~IEEE,} \\ Quan Wang, Xinbo Gao,~\IEEEmembership{Fellow,~IEEE}
\IEEEcompsocitemizethanks{%
\IEEEcompsocthanksitem Ke Li, Di Wang, Ting Wang, and Quan Wang are with the School of Computer Science and Technology, Xidian University, Xi'an 710071, China. Di Wang is also with the Interdisciplinary Institute of Artificial Intelligence, Xidian University, Xi'an, Shaanxi 710126, China. (E-mail: like0413@stu.xidian.edu.cn; wangdi@xidian.edu.cn; wangtingt@stu.xidian.edu.cn; qwang@xidian.edu.cn)
\IEEEcompsocthanksitem Yongshan Zhu is with the School of Artificial Intelligence, Xidian University, Xi'an 710071, China. (E-mail: zhu.ys@stu.xidian.edu.cn)
\IEEEcompsocthanksitem Weiping Ni is with the Northwest Institute of Nuclear Technology, Xi'an 710024, China. (E-mail: nihao\_wpni@163.com)
\IEEEcompsocthanksitem Tao Lei is with the School of Physics and Information Engineering, Fuzhou University, Fuzhou 350108, China. (E-mail: leitao@sust.edu.cn)
\IEEEcompsocthanksitem Xinbo Gao is with the Interdisciplinary Institute of Artificial Intelligence, Xidian University, Xi'an, Shaanxi 710126, China. (E-mail: xbgao@mail.xidian.edu.cn)
\IEEEcompsocthanksitem Di Wang is the corresponding author.}}

\IEEEtitleabstractindextext{%
\begin{abstract} 
\justifying{
Remote sensing visual grounding (RSVG) aims to localize a referred target in a remote sensing image or video according to a natural language expression.
Existing RSVG methods usually rely on task-specific manual annotations, which are costly to collect and inevitably limited in covering the diversity of real-world geospatial scenarios. 
As a result, they often struggle to generalize to open-vocabulary queries involving novel objects, fine-grained attributes, complex spatial relationships, and functional semantics.
In this paper, we propose RSVG-ZeroOV, a training-free framework that leverages frozen generic foundation models for zero-shot open-vocabulary RSVG.
RSVG-ZeroOV follows an \textit{Overview-Focus-Evolve} paradigm, which exploits the distinct yet complementary attention patterns of vision-language models (VLMs) and diffusion models (DMs) to progressively generate precise grounding results.
Specifically, 
(i) \textit{Overview} utilizes a VLM to extract cross-attention maps that capture semantic correlations between the referring expression and visual regions; 
(ii) \textit{Focus} leverages the fine-grained modeling priors of a DM to compensate for object structure and shape information often overlooked by VLM attention; 
and (iii) \textit{Evolve} introduces a simple yet effective attention evolution module to suppress irrelevant activations, yielding purified object masks.
To handle video inputs, we further present Video RSVG-ZeroOV, which extends image-level grounding to spatio-temporal grounding through a query-relevant key-frame selector and a temporal propagator, enabling efficient and temporally coherent video grounding without video annotations or fine-tuning.
Extensive experiments on six image and video grounding benchmarks show that RSVG-ZeroOV consistently outperforms existing zero-shot baselines and achieves competitive or superior performance compared with weakly- and fully-supervised methods.
\begin{IEEEkeywords}
Visual grounding, Open-vocabulary, Remote sensing image and video, Vision-language model
\end{IEEEkeywords}
}
\end{abstract}
}

\maketitle

\IEEEdisplaynontitleabstractindextext
\IEEEpeerreviewmaketitle

\begin{figure*}[t]
\centering
\includegraphics[width=\linewidth]{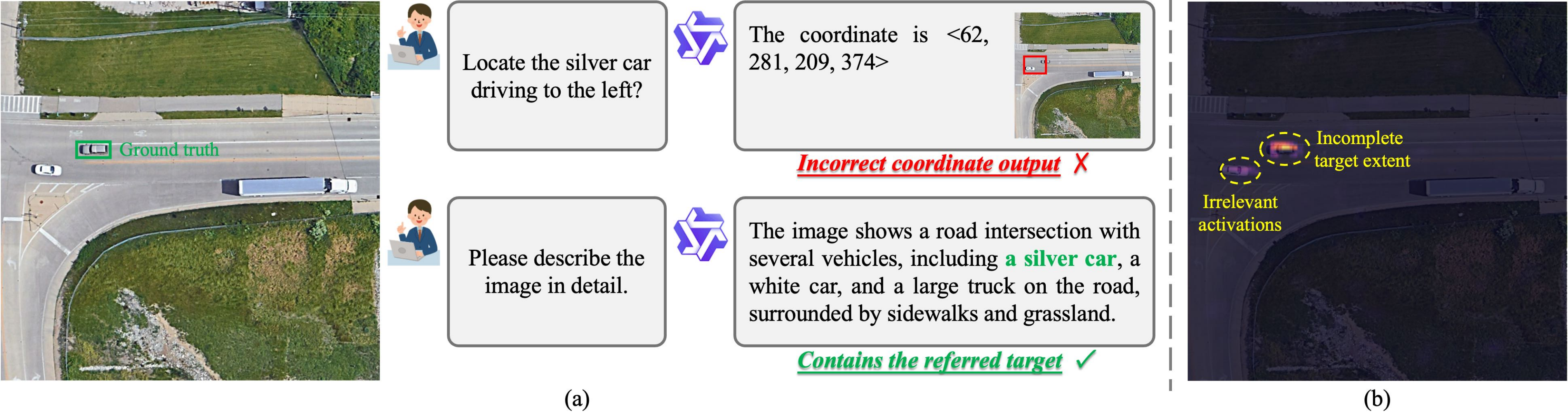}
\caption{Motivation of the proposed training-free open-vocabulary RSVG framework. 
(a) Directly prompting a generic VLM to output target coordinates may lead to inaccurate localization, while the same model can correctly describe the referred target and its surrounding context through image captioning. 
(b) The VLM cross-attention map highlights regions related to the referred target, indicating that VLMs contain useful open-vocabulary semantic cues but lack reliable coordinate-level grounding ability. 
This motivates us to convert the internal attention cues of frozen foundation models into accurate grounding masks.}
\label{fig:teaser}
\end{figure*}

\section{Introduction}
\IEEEPARstart{G}{iven} a remote sensing image or video and a natural language expression, remote sensing visual grounding (RSVG) aims to localize the referred target with a bounding box or a pixel-wise segmentation mask.
By allowing users to refer to targets in geospatial scenes through natural language, RSVG provides an intuitive interface for complex remote sensing data and supports a wide range of applications, such as disaster assessment, urban planning, UAV-based surveillance, traffic monitoring, and interactive remote sensing image/video analysis~\cite{ke2026pixdlm,schedl2021autonomous,yao2026remotereasoner,diao2025ringmo}.

Despite its practical value, extending RSVG toward real-world scenarios remains challenging.
Remote sensing data collection itself is inherently difficult, as images and videos need to cover broad geographic regions, large variations in object scale, complex spatial layouts, and dynamic temporal contexts.
The annotation burden is further amplified in RSVG, where each sample requires not only accurate localization labels (\textit{e.g.}, bounding boxes or segmentation masks) but also natural language expressions aligned with the referred target.
More importantly, existing RSVG annotations can only cover a limited range of object categories, visual concepts, and expression patterns.
Open-vocabulary RSVG goes beyond this constrained setting by allowing users to refer to targets from unseen categories with more flexible and diverse expressions, such as visual attributes, spatial relationships, functional roles, and task-specific intentions~\cite{li2026rsvg}.
As a result, models trained on such limited annotations may struggle to generalize to novel targets and diverse user queries, making them less effective in open-vocabulary RSVG.

Recent generic foundation models have demonstrated remarkable open-vocabulary perception capabilities~\cite{liu2023visual,kirillov2023segment,liu2024grounding,lin2024training}. 
Benefiting from large-scale image-text pre-training, vision-language models (VLMs) can effectively capture semantic relationships between textual descriptions and visual elements, thereby achieving impressive vision-language alignment.
Such capabilities make frozen VLMs promising for reducing the dependence on exhaustive task-specific annotations and extending RSVG toward open-vocabulary scenarios.

However, directly applying generic VLMs to open-vocabulary RSVG in a zero-shot manner remains unreliable. 
As shown in Figure~\ref{fig:teaser} \textcolor{red}{(a)}, when prompted to localize the referred target with coordinates, a VLM may produce inaccurate bounding boxes.
Nevertheless, the same model can generate a detailed caption that correctly mentions the target and its surrounding context, suggesting that useful semantic information has already been perceived.
To inspect this behavior, we visualize the VLM cross-attention maps\footnote{For decoder-only VLMs, we refer to the image-text interaction within self-attention as cross-attention to distinguish it from pure visual self-attention.} during coordinate prediction.
As shown in Figure~\ref{fig:teaser} \textcolor{red}{(b)}, although the predicted coordinates are inaccurate, the attention maps still respond to target-related regions. 
These cues, however, are insufficient for precise grounding: they often focus on object boundaries, corners, or discriminative parts rather than the complete target extent, and may include scattered activations from irrelevant regions.
Therefore, the key challenge is how to transform useful yet imperfect attention cues into accurate and complete grounding results.

These observations naturally lead to a question: \textit{Can frozen generic foundation models be effectively leveraged for zero-shot open-vocabulary grounding in remote sensing images and videos, without task-specific training?}
To answer this, we conduct a series of exploratory experiments to progressively analyze the impact of different foundation models and architectural components on performance, as detailed in Section~\ref{sec:guide}.
Through these explorations, we summarize three empirical guidelines:
\textbf{1)} Generic vision-language models exhibit strong generalization ability, which is critical for object localization in remote sensing.
\textbf{2)} Compared with other visual foundation models, diffusion models (DMs) provide stronger cues for object structure and spatial extent.
\textbf{3)} Attention maps from VLMs and DMs are complementary, and their integration consistently improves grounding performance.

Based on these guidelines, we propose \textbf{RSVG-ZeroOV}, a training-free framework for zero-shot open-vocabulary RSVG.
Our framework follows an \textit{Overview-Focus-Evolve} strategy that leverages the distinct yet complementary attention patterns of frozen VLM and DM to form the basis of our grounding results.
Specifically, RSVG-ZeroOV first aggregates cross-attention maps from all transformer heads in the VLM to obtain an initial overview of regions related to the referred target.
It then introduces an attention interaction module to integrate the cross-attention maps with self-attention maps, allowing the initial activation to better expand from sparse discriminative regions to the complete target extent.
Finally, an attention evolution module progressively suppresses irrelevant activations and refines the grounding result into an accurate and purified segmentation mask.
To extend RSVG-ZeroOV from images to videos, we further develop \textbf{Video RSVG-ZeroOV}, which introduces a query-relevant key-frame selector and a temporal propagator based on SAM3~\cite{carion2025sam}.
The key-frame selector identifies a representative frame according to its semantic similarity with the expression, on which RSVG-ZeroOV provides an initial query-aware mask.
The temporal propagator then extends the key-frame mask to the whole video, enabling training-free spatio-temporal grounding without video-level annotations or task-specific fine-tuning.

\begin{figure}[!t]
    \centering
    \includegraphics[width=\linewidth]{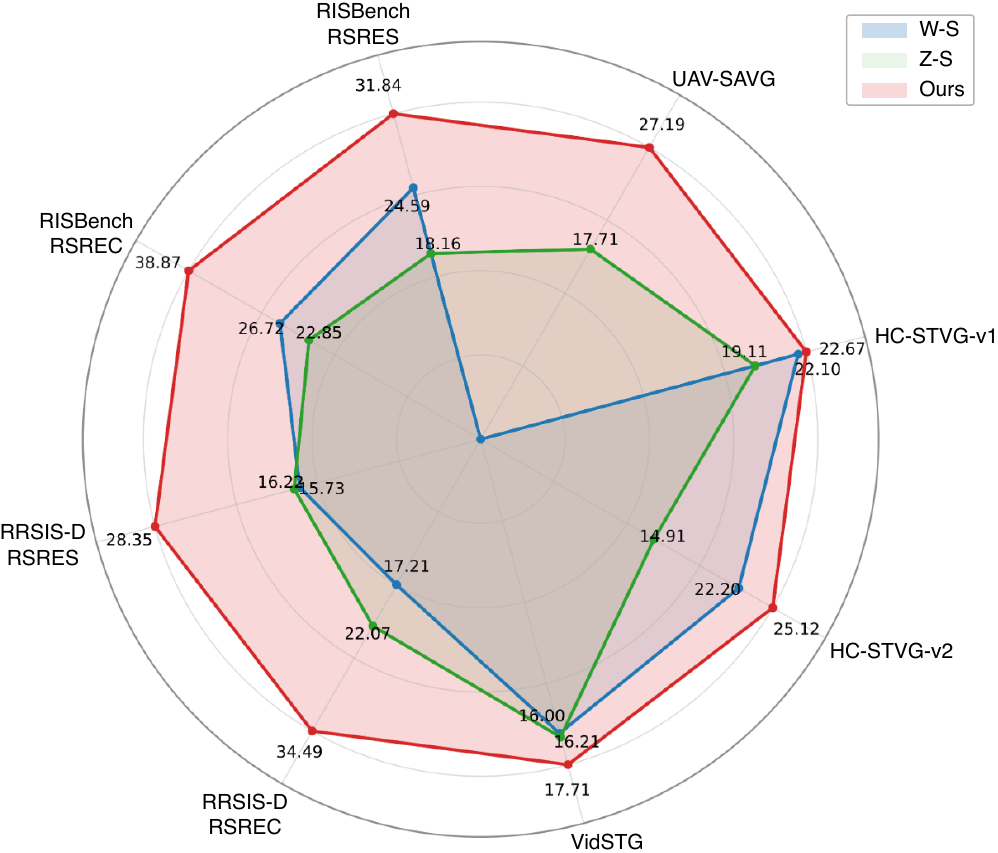}
    \caption{Overall comparison between weakly-supervised (W-S) models, zero-shot (Z-S) models, and our RSVG-ZeroOV across different tasks. For each task, we report the best result among all W-S and Z-S models, respectively.}
    \label{fig:teaser_exp}
\end{figure}

To evaluate the effectiveness and generalization ability of the proposed framework, we conduct extensive experiments across diverse grounding settings, including remote sensing image grounding on RRSIS-D~\cite{liu2024rotated} and RISBench~\cite{dong2024cross}, remote sensing video grounding on UAV-SAVG~\cite{zhan2025does}, and general-domain spatio-temporal grounding on HC-STVG-v1/-v2~\cite{tang2021human} and VidSTG~\cite{zhang2020does}.
As summarized in Figure~\ref{fig:teaser_exp}, RSVG-ZeroOV consistently outperforms existing zero-shot baselines across diverse grounding tasks and even achieves superior results compared with weakly supervised methods.
In summary, our contributions are as follows:
\begin{itemize}
  \item We propose RSVG-ZeroOV, a training-free framework for zero-shot open-vocabulary RSVG, which localizes user-intended targets from free-form language expressions without task-specific annotations or fine-tuning.
  \item We introduce an \textit{Overview-Focus-Evolve} strategy that exploits frozen generic foundation models by integrating VLM and DM attention maps to enhance the semantic and structural perception of referred objects.
  \item We extend RSVG-ZeroOV from image-level grounding to video-level grounding by introducing two simple yet effective components, \emph{i.e.,} a key-frame selector and a temporal propagator.
  \item We obtain promisingly competitive results on two benchmark datasets for image grounding and four benchmark datasets for video grounding, which demonstrates the effectiveness and generality of the proposed methods.
\end{itemize}

It is worth noting that a preliminary conference version of this work appeared in~\cite{li2026rsvg}. 
This journal extension substantially broadens the original scope from image-based grounding to video-based spatio-temporal grounding by introducing Video RSVG-ZeroOV. 
Specifically, Video RSVG-ZeroOV extends RSVG-ZeroOV to remote sensing videos through a query-relevant key-frame selector and a SAM3-based temporal propagator, without requiring video-level annotations or task-specific fine-tuning. 
To the best of our knowledge, this is among the first attempts to investigate training-free zero-shot open-vocabulary visual grounding for remote sensing in both image and video domains using frozen generic foundation models. 
Beyond remote sensing scenarios, we further evaluate the proposed framework on general-domain spatio-temporal grounding benchmarks to examine its cross-domain generalization ability. 
Compared with the conference version, this extension provides substantial new materials, including the video grounding framework, additional remote sensing and general-domain video experiments, more comprehensive ablation studies, visualization results, and in-depth analyses.

\begin{figure*}[!t]
    \centering
    \includegraphics[width=\linewidth]{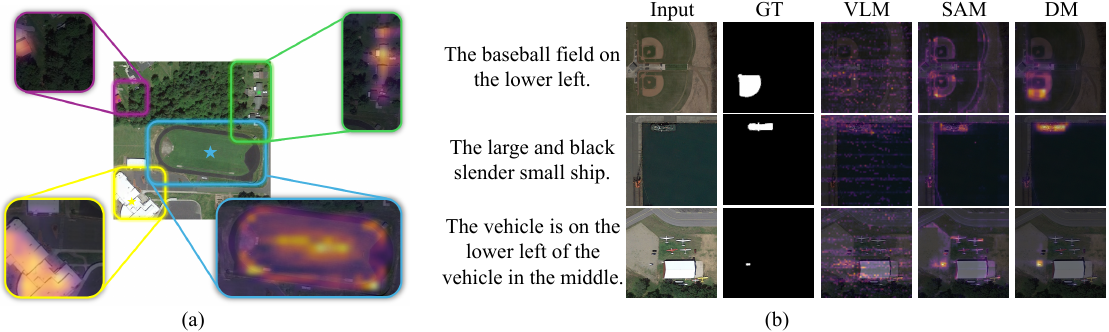}
    \caption{(a) Visualization of receptive fields derived from self-attention maps in DM.  
    (b) Comparison of attention embedding results using self-attention maps from different models.}
    \label{fig:guide2}
\end{figure*}


\section{Related Work}
\label{sec:related_work}
We review the literature most relevant to our work, including generic foundation models, remote sensing visual grounding, and open-vocabulary visual perception.

\subsection{Generic Foundation Models.}
The emergence of large-scale pretraining has fueled the rapid advancement of multimodal foundation models. 
Among them, vision-language models (VLMs) are designed for cross-modal understanding by aligning visual and textual representations. 
Early attempts, represented by CLIP~\cite{radford2021learning}, demonstrated impressive zero-shot generalization by contrastively pretraining on massive image-text pairs. 
LLaVA~\cite{liu2023visual} builds upon CLIP by adding a language decoder Vicuna~\cite{chiang2023vicuna} to support multimodal instruction-following.
More recent approaches, such as CogVLM~\cite{wang2024cogvlm}, InternVL~\cite{chen2024internvl}, and Qwen-VL~\cite{wang2024qwen2}, further enhance this paradigm and demonstrate outstanding performance in open-world visual understanding.
However, due to their emphasis on high-level semantic reasoning and lack of pixel-level supervision, these models often struggle with fine-grained spatial perception.
Another important branch is diffusion models (DMs)~\cite{rombach2022high}, which focus on visual content generation from text prompts.
Recent studies~\cite{pnvr2023ld,xu2023open} reveal that the hierarchical attention maps in DMs encode strong structural priors, making them useful for downstream tasks such as classification, detection, and segmentation.
In addition, segmentation foundation models, represented by SAM~\cite{kirillov2023segment} and its video extensions such as SAM2~\cite{ravi2025sam} and SAM3~\cite{carion2025sam}, provide promptable mask generation and temporal propagation capabilities.
In this study, we focus on exploiting the complementary roles of VLMs and DMs for query-aware grounding, while using SAM-based models as plug-and-play tools for optional mask refinement and temporal propagation.

\subsection{Remote Sensing Visual Grounding.} 
\subsubsection{Image-Based RSVG}
Image-based RSVG aims to localize target objects in static remote sensing imagery based on natural language descriptions. 
Compared with natural scene images, remote sensing images often exhibit large-scale variations, complex backgrounds, dense small objects, and diverse imaging conditions. 
Remote sensing referring expression comprehension (RSREC) and segmentation (RSRES) are two typical image-based RSVG tasks. 
RSREC aims to predict a bounding box corresponding to the given referring expression, while RSRES requires a pixel-wise binary mask of the referred object. 
The pioneering work~\cite{sun2022visual} explores RSREC with a one-stage dense prediction framework. 
Recently, transformer-based methods~\cite{zhan2023rsvg,li2024language,li2026provg,lan2024language,ding2024visual} have been introduced to better capture cross-modal context and improve localization accuracy. 
For RSRES, recent works~\cite{yuan2023rrsis,liu2024rotated,dong2024cross,li2025segearth_r1} improve segmentation quality by enhancing multi-scale visual-text interactions. 
Despite these advances, existing image-based RSVG methods still heavily rely on task-specific annotations and have limited generalization to open-vocabulary expressions.
\subsubsection{Video-Based RSVG}
Video-based RSVG extends visual grounding from static images to temporal sequences, requiring consistent localization of the referred target across frames. 
Compared with image-based settings, it introduces additional challenges such as object motion, camera motion, occlusion, scale variation, and temporal changes in object appearance or spatial relations. 
Due to the limited availability of remote sensing video grounding datasets, this task remains relatively underexplored.
Existing studies are mainly developed in the general video domain, including video referring expression comprehension and spatio-temporal video grounding. 
Early works often follow tracking- or proposal-based pipelines to generate phrase-relevant candidate regions and associate them across frames~\cite{feng2020real,feng2021siamese}. 
Recent fully-supervised methods adopt detection-based or transformer-based architectures to better model temporal dependencies and cross-modal interactions~\cite{song2021co,cao2022correspondence,yang2024language,yang2022tubedetr,jin2022embracing,wang2023efficient,gu2024context,wasim2024videogrounding}. 
Beyond fully-supervised learning, some recent works explore weakly-supervised~\cite{li2023winner,jin2024weakly,kumar2025contextual,garg2025stpro} or zero-shot video grounding~\cite{subramanian2022reclip,shtedritski2023does,bao2024e3m} to reduce the dependence on dense video annotations. 
Inspired by these advances, SAVG-DETR~\cite{zhan2025does} makes an initial attempt to address aerial video grounding in remote sensing scenarios. 
Nevertheless, existing remote sensing video grounding methods still rely on task-specific video annotations and provide limited support for zero-shot open-vocabulary grounding. 
This motivates us to explore a training-free framework that extends image-level open-vocabulary RSVG to spatio-temporal grounding in remote sensing videos.

\subsection{Open-Vocabulary Visual Perception} 
With the rapid development of multimodal foundation models, open-vocabulary learning has achieved remarkable success in natural image domains~\cite{wu2024toward,han2022expanding,lin2026vl}. 
Inspired by this progress, recent studies have extended these paradigms to the remote sensing field~\cite{wei2026mm,li2026exploiting,li2024toward}.
For instance, RemoteCLIP~\cite{liu2024remoteclip} improves transferability in downstream tasks by pretraining on extensive remote sensing image-text pairs. 
Similarly, VisGT~\cite{pan2025locate} and GSNet~\cite{ye2025towards} construct large-scale open-vocabulary detection and segmentation datasets, respectively, enabling models to recognize a broader range of object categories.
However, building high-quality annotated datasets for remote sensing remains costly and labor-intensive.
To alleviate the reliance on manual annotations, several efforts~\cite{zheng2026instructsam} have explored training-free approaches.
For example, SegEarth-OV~\cite{li2025segearth_ov,li2025annotation} adapts existing generic models by modifying or removing task-specific components, allowing seamless application to high-resolution remote sensing segmentation.
Nevertheless, these methods still rely on predefined category labels during inference stage and lack the flexibility to understand user-intended objects described in free-form language.
As discussed in the introduction, users in practical applications often prefer to locate objects by describing visual attributes, spatial relationships, or functional roles using natural language expressions.
To the best of our knowledge, this work is the first to investigate zero-shot open-vocabulary RSVG, which requires models to flexibly interpret free-form expressions and localize the referred objects without task-specific training.
To achieve this goal, we propose RSVG-ZeroOV, a training-free framework that enables frozen foundation models to perform expression-driven open-vocabulary grounding beyond predefined category labels.

\begin{table}[!t]
\centering
\caption{Comparison of generic VLMs and remote sensing VLM on RRSIS-D test set, all using 7B-scale LLMs.}
\resizebox{\linewidth}{!}{
\begin{tabular}{l|cc|cc}
    \specialrule{.1em}{.3em}{.3em} 
    \multicolumn{1}{c|}{}                          & \multicolumn{2}{c|}{RSREC} & \multicolumn{2}{c}{RSRES}\\
    \cmidrule(r){2-5}       
    \multicolumn{1}{l|}{Method (Zero-Shot)}        & Pr@0.5 & mIoU & Pr@0.5 & mIoU\\
    \specialrule{.1em}{.3em}{.3em} 
    GeoChat     (CVPR'24)                          & 23.93 & 30.12 & -     & - \\
    LLaVA-1.5   (CVPR'24)                          & 13.21 & 20.59 & -     & - \\
    Qwen2.5-VL  (arXiv'25)                         & 28.66 & 30.90 & -     & - \\
    \midrule
    GeoChat + SAM                                  & 27.61 & 32.53 & 18.85 & 24.92 \\
    LLaVA-1.5 + SAM                                & 16.26 & 21.97 & 10.86 & 16.86 \\
    Qwen2.5-VL + SAM                               & \textbf{30.35} & \textbf{31.93} & \textbf{24.68} & \textbf{25.72} \\
    \specialrule{.1em}{.3em}{.3em} 
\end{tabular}}
\label{tab:guide1}
\end{table}

\section{Guidelines of Exploring Generic Foundation Models}
\label{sec:guide}
Naively transferring generic foundation models to open-vocabulary remote sensing visual perception tasks does not yield satisfactory results.  
In this section, we summarize three guidelines for the effective use of foundation models.

\noindent \textbf{Guideline 1: generic VLMs exhibit strong generalization ability on unseen remote sensing scenes.}
To demonstrate this, we conduct zero-shot experiments on RRSIS-D test set. 
Specifically, we compare three 7B-scale models, including LLaVA-1.5~\cite{liu2024improved}, Qwen2.5-VL~\cite{Qwen2.5-VL}, and GeoChat~\cite{kuckreja2024geochat}. 
For RSREC task, we generate bounding box predictions of referred objects using a unified prompt format: [Locate the object referred to by ‘\{referring expression\}’ and return its box coordinates (x1, y1, x2, y2)].
Notably, GeoChat has been finetuned on 318k image-instruction pairs collected from remote sensing datasets, enabling it to learn more domain-specific priors.
However, as shown in Table~\ref{tab:guide1}, GeoChat achieves 23.93\% Pr@0.5 on RSREC task, noticeably lower than Qwen2.5-VL's \textbf{28.66\%}. 
This result suggests that VLMs pretrained on large-scale generic datasets can exhibit strong cross-domain generalization even without access to remote sensing data.
For RSRES task, we utilize outputs of VLMs as box prompts for SAM to get pixel-level masks.
As we can see, Qwen2.5-VL still outperforms GeoChat by \textbf{0.8\%} in mIoU.
These findings highlight the potential of generic VLMs as a solid foundation for zero-shot open-vocabulary RSVG, even without any domain-specific training.

\noindent \textbf{Guideline 2: self-attention maps in DM encode superior structural priors of objects.}
Theoretically, self-attention mechanisms are inherently suited for modeling object structure, as they compute pairwise similarities across visual elements, enabling the network to capture spatial dependencies and shape layouts.
As shown in Figure~\ref{fig:guide2} \textcolor{red}{(a)}, we visualize the receptive fields of self-attention maps anchored at different query points. 
The results demonstrate that self-attention can effectively distinguish foreground objects from the background, reflecting a strong ability for structural perception.
To further investigate this, we compare the spatial distributions of self-attention extracted from three representative architectures: the visual encoder of VLM, the image backbone of SAM, and the U-Net of DM, as illustrated in Figure~\ref{fig:guide2} \textcolor{red}{(b)}. 
Specifically, the VLM attention maps tend to be broadly distributed and spatially scattered, consistent with its objective of capturing global semantic context.
While SAM produces sharper boundaries, its purely visual design lacks semantic understanding, which often leads to over-attention on surrounding background areas.
For instance, in the second row, SAM simultaneously highlights both the ship and port. 
In contrast, the self-attention maps from DM display the most coherent structural representations, \textit{i.e.,} attention is evenly and densely distributed across the entire object extent, revealing a more precise and holistic understanding of object shapes.
To quantitatively validate these observations, we integrate the different types of self-attention maps into our visual grounding pipeline and evaluate their performance.
As reported in Table~\ref{tab:guide23}, the use of self-attention maps from DM (the last row) achieves the highest Pr@0.5 and mIoU on both RSREC and RSRES benchmarks.
These results confirm the superiority of DM's self-attention in capturing object-centric structural cues for visual grounding tasks.

\begin{table}[!t]
\centering
\caption{Evaluation of different self-attention maps on RRSIS-D test set.}
\resizebox{\linewidth}{!}{
\begin{tabular}{l|cc|cc}
    \specialrule{.1em}{.3em}{.3em} 
    \multicolumn{1}{c|}{}                         & \multicolumn{2}{c|}{RSREC} & \multicolumn{2}{c}{RSRES}\\
    \cmidrule(r){2-5}       
    \multicolumn{1}{l|}{Self-Attention Maps} & Pr@0.5 & mIoU & Pr@0.5 & mIoU\\
    \specialrule{.1em}{.3em}{.3em}
    \textit{w/o} Cross-Attention Map                     & 21.49 & 26.26 & 1.18  & 6.15 \\
    ViT in VLM                       & 12.53 & 21.02 & 2.13  & 7.40\\
    ViT in SAM                       & 18.85 & 23.56 & 4.77  & 10.23 \\
    U-Net in DM (Ours)               & \textbf{30.15} & \textbf{32.92} & \textbf{12.84} & \textbf{21.85}   \\
    \specialrule{.1em}{.3em}{.3em} 
\end{tabular}}
\label{tab:guide23}
\end{table}

\noindent \textbf{Guideline 3: combining cross- and self-attention maps is vital especially for pixel-level perception tasks.}
To validate this, we first evaluate the grounding performance using only the cross-attention maps from the VLM, as shown in Table~\ref{tab:guide23} (the first row). 
Removing the self-attention maps from DM leads to significant performance drops of \textbf{8.66\%} Pr@0.5 and \textbf{6.66\%} mIoU on RSREC, and \textbf{11.66\%} and \textbf{15.70\%} on RSRES, respectively.
Such a phenomenon indicates the necessity of combining cross- and self-attention maps, especially for pixel-level perception tasks like RSRES.
These findings naturally lead to a question: \textit{which interaction strategy best integrates cross- and self-attention maps for optimal grounding performance?}
We try four strategies: 
1) Anchor-based: selecting high-response pixels from the cross-attention map to represent the object, followed by aggregating their corresponding self-attention regions.
2) Multiplication: performing element-wise multiplication on cross- and self-attention maps, denoted as $\mathcal{A}_S \cdot \mathcal{A}_C$. 
3) Exponentiation: enhancing the self-attention map by raising it to a power $\gamma$, \textit{i.e.}, $\mathcal{A}_S^{\gamma} \cdot \mathcal{A}_C$. 
4) Similarity: computing cosine similarity between the cross- and self-attention maps at each spatial location (details in the next section).
As summarized in Table~\ref{tab:guide3}, although the anchor-based and exponentiation strategies achieve reasonable performance on RSREC, they suffer on RSRES. The former tends to oversimplify by omitting many regions, while the latter tends to overamplify high activations. 
In contrast, the proposed similarity strategy yields more semantically consistent initial masks, which allows the evolve stage to further improve segmentation quality, ultimately achieving a better performance on both benchmarks.

\begin{table}[!t]
\centering
\caption{Comparison of different interaction strategies between cross- and self-attention maps on RRSIS-D test set.}
\resizebox{\linewidth}{!}{
\begin{tabular}{l|cc|cc}
    \specialrule{.1em}{.3em}{.3em} 
    \multicolumn{1}{c|}{}                         & \multicolumn{2}{c|}{RSREC} & \multicolumn{2}{c}{RSRES}\\
    \cmidrule(r){2-5}       
    \multicolumn{1}{l|}{Interaction Strategy} & Pr@0.5 & mIoU & Pr@0.5 & mIoU\\
    \specialrule{.1em}{.3em}{.3em}
    Anchor-based                        & 23.18 & 26.94 & 6.58  & 16.69 \\
    \ \ + \textit{Evolve stage}         & 28.73 & 31.48 & 7.61  & 16.58 \\
    \midrule
    Multiplication                      & 21.94 & 26.32 & 10.03 & 19.78 \\
    \ \ + \textit{Evolve stage}         & 29.26 & 31.54 & 11.52 & 20.75 \\
    \midrule
    Exponentiation                      & 24.07 & 28.61 & 5.37  & 15.30 \\
    \ \ + \textit{Evolve stage}         & 27.38 & 31.02 & 5.31  & 14.00 \\
    \midrule
    \textit{Similarity} (Ours)          & 22.63 & 26.65 & 10.26 & 20.56 \\
    \ \ + \textit{Evolve stage}         & \textbf{30.15} & \textbf{32.92} & \textbf{12.84} & \textbf{21.85}   \\
    \specialrule{.1em}{.3em}{.3em} 
\end{tabular}}
\label{tab:guide3}
\end{table}

\begin{figure*}[!t]
\centering
\includegraphics[width=\linewidth]{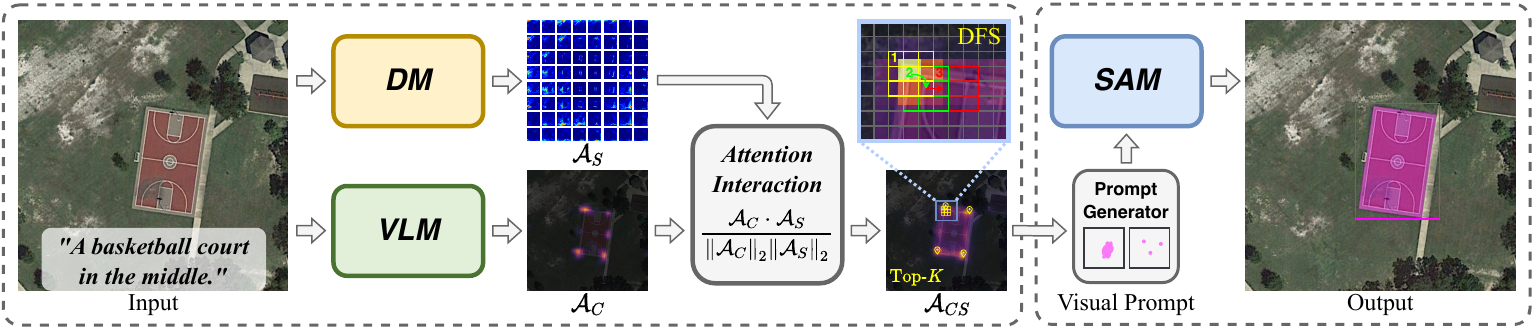}
\caption{
The overall pipeline of the proposed RSVG-ZeroOV for image-based RSVG. 
The VLM branch extracts language-aware cross-attention maps $\mathcal{A}_C$ for coarse target localization, while the DM branch provides self-attention maps $\mathcal{A}_S$ as object-structure priors. 
Then, $\mathcal{A}_C$ and $\mathcal{A}_S$ are interacted to generate the structurally enhanced attention map $\mathcal{A}_{CS}$, from which top-$K$ confident seeds are selected and isolated noisy responses are removed by depth-first search (DFS)-based connected-region filtering.
Finally, the denoised connected regions are converted into visual prompts and fed into SAM to produce the final segmentation mask and bounding box.}
\label{fig:framework_image}
\end{figure*}

\section{Methodology}
\label{sec:rsvg_zeroov}
\begin{figure*}[!t]
\centering
\includegraphics[width=\linewidth]{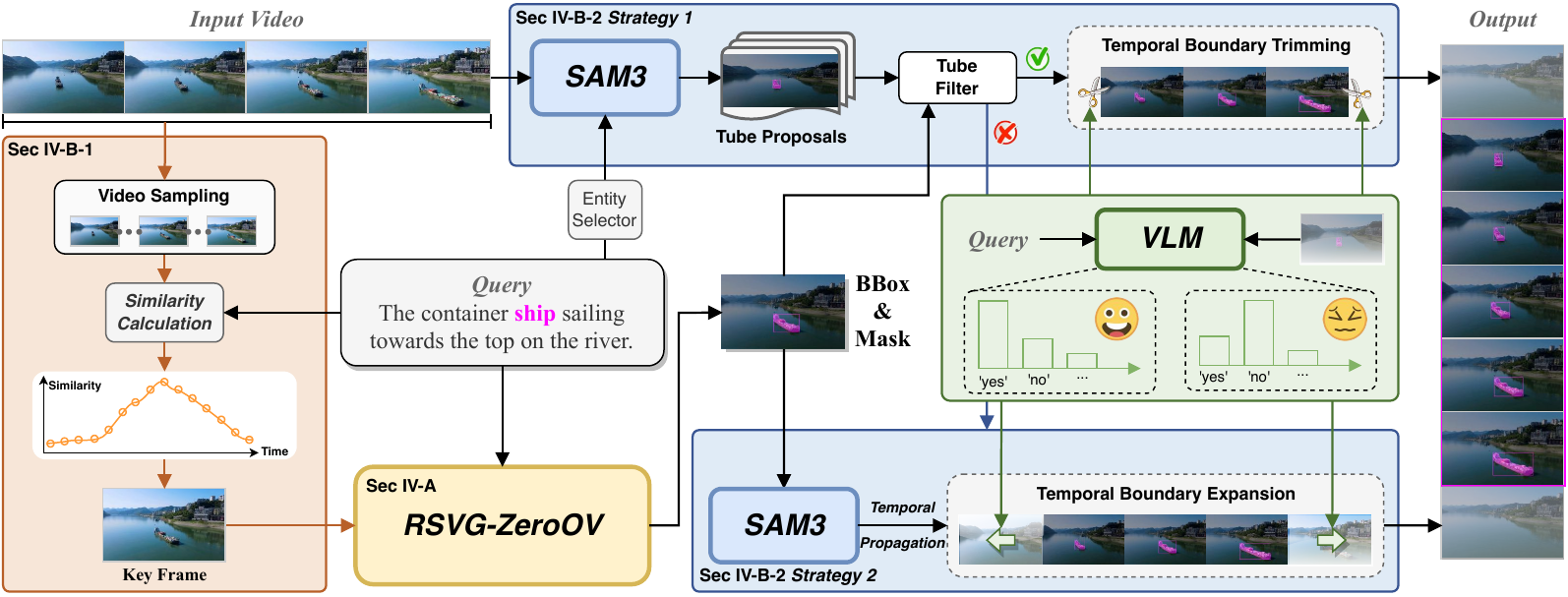}
\caption{
The overall pipeline of the proposed Video RSVG-ZeroOV for video-based RSVG. 
A key frame is first selected from uniformly sampled frames by measuring the similarity between each frame and the referring expression. 
The image-level RSVG-ZeroOV produces the key-frame grounding result, including a bounding box and mask, which is used as the query-aware spatial anchor. 
In Strategy 1, SAM3 generates tube proposals with the extracted entity prompt, and a tube filter selects the proposal that best matches the key-frame result, followed by VLM-based temporal boundary trimming. 
In Strategy 2, SAM3 takes the key-frame mask as a visual prompt and performs bidirectional temporal propagation, followed by VLM-based temporal boundary expansion and verification.} 
\label{fig:framework_video}
\end{figure*}

In this section, we start with the introduction of the proposed RSVG-ZeroOV framework for image-based RSVG (illustrated in Figure~\ref{fig:framework_image}) in Section~\ref{sec:image_grounding_method}.
Then, we describe how to extend the framework to video-based RSVG (illustrated in Figure~\ref{fig:framework_video}) in Section~\ref{sec:video_grounding_method}.

\subsection{RSVG-ZeroOV for Image Grounding}
\label{sec:image_grounding_method}
Following our proposed guidelines, we present a straightforward training-free framework named \textbf{RSVG-ZeroOV} for zero-shot open-vocabulary RSVG.
As illustrated in Figure~\ref{fig:framework_image}, the framework comprises three sequential stages:
\textit{(i) Overview:} we prompt a VLM with a referring expression to identify the referred object and extract its corresponding cross-attention map, which highlights regions most relevant to the given textual query.
\textit{(ii) Focus:} we cache the self-attention maps from the U-Net of a DM to obtain structural priors of objects, and then propose an effective interaction strategy to embed these priors into the cross-attention distribution.
\textit{(iii) Evolve:} to generate a high-quality attention map, we introduce an attention evolution module to filter out irrelevant activations, thus improving the quality of the segmentation mask.

\begin{figure}[!t]
    \centering
    \includegraphics[width=\linewidth]{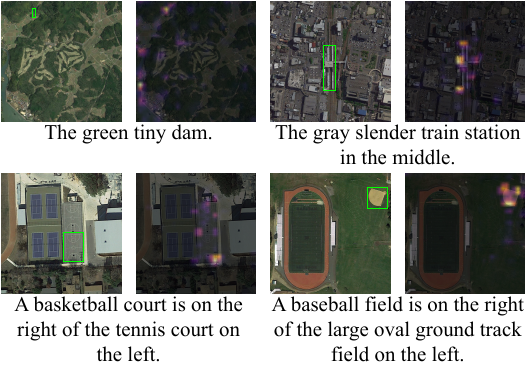}
    \caption{Visualization of some cross-attention maps from VLM on RRSIS-D.}
    \label{fig:guide1weak}
\end{figure}

\textbf{Overview Stage.} 
Referring to \textbf{Guideline 1}, we utilize the attention map from a frozen VLM to capture the coarse-level localization cues of the referred object.
Given a remote sensing image $\mathcal{I}$ and a referring expression $\mathcal{T}$, we first hook into the attention weights $\mathcal{W}^{(t)} \in \mathbb{R}^{H \times 1 \times N}$ from VLM, where $N$ is the token length, $H$ is the number of attention heads, and $t \in \{1, \dots, T\}$ indexes the $T$ autoregressive forward passes.
To obtain cross-attention parts, we extract the image-text-related segment $\mathcal{W}^{(t)}_{p:p'}$, where $p$ and $p'$ indicate the span of visual tokens in the input sequence.
After that, we aggregate information from the transformer heads with mean weights, and form a sentence-level cross-attention map by averaging all $t$ as:
\begin{equation}
  \label{eq:equ0}
  \mathcal{A}_C = 1/T {\textstyle \sum_{t=1}^{T}} ( 1/H {\textstyle \sum_{h=1}^{H}} \mathcal{W}^{(t)}_{p:p'} ).
\end{equation}
As shown in Figure~\ref{fig:guide1weak}, cross-attention maps establish the relationship between textual queries and visual pixels. 
However, it exhibits two key limitations:
\textit{(i) Attention tends to focus on object boundaries or corners rather than the full extent of the object.} 
This behavior stems from the high-level semantic concentration of VLMs, which encourages attention to concentrate on key features of objects.
\textit{(ii) Attention is often scattered and includes irrelevant regions.} 
This dispersion arises because VLMs need to aggregate contextual cues from multiple visual regions to understand complex referring expressions.

\textbf{Focus Stage.} 
To mitigate the first \textit{issue (i)}, referring to the aforementioned \textbf{Guidelines 2\&3}, we propose an attention interaction module that integrates cross-attention with structural priors derived from self-attention. 
Considering the diverse scales of objects in remote sensing images, we first extract multi-scale self-attention maps from the U-Net in DM, and fuse them to form a unified structural prior $\mathcal{A}_S\in \mathbb{R}^{H \times W \times H \times W}$, which is denoted as:
\begin{equation}
\mathcal{A}_S = 1/L {\textstyle \sum_{l\in L}}(\mathcal{A}^{l}_S),
\label{eq:equ1}
\end{equation}
where $\mathcal{A}_S^l$ denotes the self-attention map from $l$-th layer.
Next, we calculate the correlation score\footnote[3]{In fact, $\mathcal{A}_C$ does not match the spatial resolution of $\mathcal{A}_S$, hence we interpolate $\mathcal{A}_C$ to the same size as $\mathcal{A}_S$ before interaction.} between $\mathcal{A}_C$ and $\mathcal{A}_S$ through cosine similarity:
\begin{equation}
  \label{eq:equ2}
  \mathcal{A}_{CS} = \frac{\mathcal{A}_C \cdot \mathcal{A}_S}{\|\mathcal{A}_C\|_2 \|\mathcal{A}_S\|_2},
\end{equation}
where $\|\cdot\|_2$ denotes the $L_2$ norm. 
This interaction introduces object-centric structural guidance into the original cross-attention map, allowing the model to shift focus from incomplete boundaries to semantically coherent regions.

\textbf{Evolve Stage.} 
Although this interaction strategy remarkably fills the gaps within the object area, it also exacerbates the impact of \textit{issue (ii)}.
To address this, we introduce an attention evolution module based on recursive region expansion to suppress scattered activation signals that are outside the referred object area.
Specifically, we first select the top-$K$ pixels from the interaction map $\mathcal{A}_{C}$ with the highest response values:
\begin{equation}
    \label{eq:equ3}
    \mathcal{S} = \mathrm{TopK}(\mathcal{A}_{C}, K),
\end{equation}
where $\mathcal{S} = \{(i_k, j_k)\}_{k=1}^{K}$ denotes the set of starting seed positions.
For each seed $(i_k, j_k)$, we define a local neighborhood kernel $\mathcal{N}(i, j)$ of size 3$\times$3, and iteratively expand the region by selecting the neighbor with the highest response:
\begin{equation}
    \label{eq:equ4}
    (i^*, j^*) =\underset{(u,v) \in \mathcal{N}(i,j)}{\arg\max} \mathcal{A}_{CS}[u, v].
\end{equation}
We then perform a depth-first search (DFS) starting from each seed to recursively grow the region. 
A pixel $(u,v)$ is included in the evolved attention map $\mathcal{A}_E$ if it satisfies:
\begin{equation}
\label{eq:equ5}
\mathcal{A}_{CS}[u, v] \geq \tau,\ \text{\&\&} \ (u,v) \in \mathrm{DFS}(\mathcal{S}),
\end{equation}
where $\tau$ is a response threshold that controls the expansion boundary, and the process continues until no further qualified neighbors can be added.
Finally, we perform binarization on the evolved attention map $\mathcal{A}_E$ to obtain the segmentation mask $\boldsymbol{M} \in \{0, 1\}^{H \times W}$, which is defined as:
\begin{equation}
    \label{eq:equ6}
    \boldsymbol{M}(i, j) = 
    \begin{cases}
    1, & \text{if } \mathcal{A}_E(i, j) \ge \alpha \\
    0, & \text{otherwise}
    \end{cases},
\end{equation}
where $\alpha \in (0, 1]$ is a predefined binarization threshold.

\textbf{Refine Stage (Optional).} 
Following prior works~\cite{liu2024rotated,dong2024cross}, we additionally introduce an optional refinement stage to further improve the segmentation mask $\boldsymbol{M}$.
We evaluate the effectiveness of several refinement methods, including DenseCRF and various prompt types for SAM.
Our results show that box prompts yield the best performance, whereas point prompts are less effective.

\subsection{RSVG-ZeroOV for Video Grounding}
\label{sec:video_grounding_method}
Given a remote sensing video $\mathcal{V}=\{\mathcal{I}_t\}_{t=1}^{L}$ and a referring expression $\mathcal{T}$ describing the target object, video-based RSVG aims to localize the referred object across frames by predicting a sequence of pixel-wise masks $\mathcal{M}=\{\boldsymbol{M}_t\}_{t=1}^{L}$. 
A naive solution is to apply image-based RSVG-ZeroOV independently to each frame. 
However, such a frame-by-frame manner is computationally inefficient and overlooks the temporal continuity of video data, which may lead to unstable predictions across adjacent frames. 
To address these issues, we propose Video RSVG-ZeroOV, a training-free extension of RSVG-ZeroOV for video grounding, as illustrated in Figure~\ref{fig:framework_video}. 

Specifically, our framework first selects a query-relevant key frame and applies image-based RSVG-ZeroOV only to this frame to obtain an initial mask of the referred object. 
The key-frame mask is then used as a query-aware spatial anchor for video-level grounding. 
Based on this anchor, we consider two training-free strategies: a proposal-based strategy that selects the target object tube from SAM3-generated candidates, and a prompt-based strategy that directly uses the key-frame mask as a visual prompt for bidirectional mask propagation with SAM3.

\subsubsection{Key-Frame Selector}
\label{sec:key_frame_selector}
To avoid processing all video frames independently, we first select a query-relevant key frame as the spatial anchor for video-level grounding.
Specifically, we uniformly sample $N$ frames from the input video:
\begin{equation}
    \mathcal{S}=\{\mathcal{I}_{t_i}\}_{i=1}^{N}, \quad t_i = \left\lfloor 1 + \frac{(i-1)(L-1)}{N-1} \right\rfloor,
    \label{eq:sample_frames}
\end{equation}
where $L$ is the total number of frames and $N$ is the number of sampled frames. 
For each sampled frame $\mathcal{I}_{t_i}$, we compute its semantic similarity with the referring expression $\mathcal{T}$ using a frozen BLIP model~\cite{li2022blip}.
Let $\phi_v(\cdot)$ and $\phi_l(\cdot)$ denote the visual and textual encoders, respectively. 
The query-frame similarity score is computed as:
\begin{equation}
    s_{t_i} = 
    \frac{\phi_v(\mathcal{I}_{t_i})^\top \phi_l(\mathcal{T})}
    {\|\phi_v(\mathcal{I}_{t_i})\|_2 \|\phi_l(\mathcal{T})\|_2}.
    \label{eq:frame_similarity}
\end{equation}
The key frame is then selected as the sampled frame with the highest semantic relevance:
\begin{equation}
    t^{*} = \arg\max_{t_i \in \mathcal{S}} s_{t_i}.
    \label{eq:key_frame}
\end{equation}
After obtaining the key frame $\mathcal{I}_{t^{*}}$, we apply the image-based RSVG-ZeroOV pipeline to segment the referred object:
\begin{equation}
    \boldsymbol{M}_{t^{*}} = \mathrm{RSVG\text{-}ZeroOV}(\mathcal{I}_{t^{*}}, \mathcal{T}),
    \label{eq:key_mask}
\end{equation}
where $\boldsymbol{M}_{t^{*}}$ denotes the predicted segmentation mask on the key frame, which serves as a query-aware spatial initialization for subsequent video-level grounding.

\subsubsection{Temporal Propagator}
\label{sec:temporal_propagation}
Given the key-frame grounding result $\boldsymbol{M}_{t^{*}}$, the remaining task is to obtain temporally coherent grounding results for the referred object across the video. 
This is non-trivial for remote sensing videos, where dense small objects, large background regions, and frequent scale variations caused by UAV motion or viewpoint changes can easily introduce drift or ambiguous object associations. 
To handle these challenges without additional training, we explore two SAM3-based strategies: proposal-based tube filtering and prompt-based temporal propagation.

\noindent \textbf{\emph{Strategy 1:} Proposal-based tube filtering.}
The proposal-based strategy first generates a set of candidate object tubes with SAM3 and then selects the tube that best matches the query-aware key-frame grounding result produced by RSVG-ZeroOV. 
Specifically, we parse the referring expression $\mathcal{T}$ to extract the target noun phrase, denoted as $\mathcal{E}$, which indicates the semantic category of the referred object. 
We then use $\mathcal{E}$ as a text prompt for SAM3 to segment potential target instances in the video. 
Since the extracted noun phrase usually provides category-level semantics rather than instance-level disambiguation, SAM3 may produce multiple candidate tubes corresponding to visually or semantically similar objects. 
Given the video $\mathcal{V}$ and the text prompt $\mathcal{E}$, SAM3 generates $K$ candidate tube proposals:
\begin{equation}
    \mathcal{P} = \{\tau_k\}_{k=1}^{K} = \mathrm{SAM3}(\mathcal{V}, \mathcal{E}),
    \label{eq:tube_proposals}
\end{equation}
where each tube proposal $\tau_k$ consists of a sequence of masks:
\begin{equation}
    \tau_k = \{{\boldsymbol{M}}_t^k\}_{t=t_s^k}^{t_e^k}.
    \label{eq:tube_sequence}
\end{equation}

To identify the tube corresponding to the referred object, we use the key-frame mask $\boldsymbol{M}_{t^{*}}$ predicted by RSVG-ZeroOV as a query-aware spatial reference. 
For each candidate tube $\tau_k$ that covers the key frame $t^{*}$, we compute the mask overlap between its key-frame mask $\boldsymbol{M}_{t^{*}}^k$ and the RSVG-ZeroOV prediction $\boldsymbol{M}_{t^{*}}$:
\begin{equation}
    o_k = 
    \frac{|\boldsymbol{M}_{t^{*}}^k \cap \boldsymbol{M}_{t^{*}}|}
    {|\boldsymbol{M}_{t^{*}}^k \cup \boldsymbol{M}_{t^{*}}|}.
    \label{eq:tube_overlap}
\end{equation}
The tube with the largest overlap is selected as the target tube:
\begin{equation}
    k^{*} = \arg\max_{k:\tau_k \in \mathcal{P}_{t^{*}}} o_k, \quad \tau^{*} = \tau_{k^{*}}.
    \label{eq:target_tube}
\end{equation}
where $\mathcal{P}_{t^{*}}=\{\tau_k \mid t^{*}\in[t_s^k,t_e^k]\}$ denotes the set of tube proposals that contain the key frame.
This matching step transfers the instance-level grounding ability of RSVG-ZeroOV to the SAM3-generated category-level tube proposals, thereby reducing ambiguity among visually similar objects.

Although the selected tube is temporally coherent, the tracked object may not satisfy the referring expression throughout the entire tube, especially when the expression involves transient attributes or spatial relations. 
We therefore trim expression-inconsistent frames using a VLM-based yes/no scoring function. 
For a frame-mask pair $(\mathcal{I}_t, {\boldsymbol{M}}_t)$ and the referring expression $\mathcal{T}$, we feed them into the frozen VLM with a binary question and extract the next-token logits of the answer tokens ``yes'' and ``no'', denoted as $\ell_1^t$ and $\ell_0^t$, respectively. 
The normalized confidence of the positive answer is defined as:
\begin{equation}
    \label{eq:binary_verification}
    \psi(\mathcal{I}_t, {\boldsymbol{M}}_t, \mathcal{T}) =
    \frac{
    \exp(\ell_1^t)
    }{
    \exp(\ell_1^t)+\exp(\ell_0^t)
    }.
\end{equation}
For the selected tube $\tau^*$, the frame-level validity indicator is:
\begin{equation}
    y_t = \mathbb{I}\left(\psi(\mathcal{I}_t, {\boldsymbol{M}}_t^{*}, \mathcal{T}) > \delta\right),
    \label{eq:tube_validity}
\end{equation}
where $\boldsymbol{M}_t^{*}$ is the mask of $\tau^{*}$ at frame $t$, and $\delta$ is the verification threshold. 
The final temporal interval is obtained by retaining the longest continuous positive segment:
\begin{equation}
    [\hat{t}_s,\hat{t}_e] =
    \mathrm{LongestContinuousSegment}
    (\{y_t\}_{t=t_s^{*}}^{t_e^{*}}).
    \label{eq:longest_segment}
\end{equation}
The final grounding result is:
\begin{equation}
    \mathcal{M} = \{{\boldsymbol{M}}_t^{*}\}_{t=\hat{t}_s}^{\hat{t}_e}.
    \label{eq:final_tube_result}
\end{equation}

\noindent \textbf{\emph{Strategy 2:} Prompt-based temporal propagation.}
The second strategy directly uses the key-frame grounding result produced by RSVG-ZeroOV as a visual prompt for SAM3. 
Unlike the proposal-based strategy, it does not rely on category-level tube proposals or target selection among multiple candidates. 
Instead, it starts from the query-aware key-frame mask and propagates the referred object along the temporal dimension.

Given the key-frame prediction $\boldsymbol{M}_{t^{*}}$, we feed the predicted mask into SAM3 as a visual prompt:
\begin{equation}
    \{{\boldsymbol{M}}_t\}_{t=1}^{L} = \mathrm{SAM3}(\mathcal{V}, {\boldsymbol{M}}_{t^{*}}, t^{*}).
    \label{eq:prompt_propagation}
\end{equation}
Starting from the key frame $t^{*}$, SAM3 propagates the prompted target both forward and backward in time, producing an initial spatio-temporal grounding sequence.

However, propagation from a single key frame may suffer from temporal over-expansion, where masks are propagated to frames in which the referred object is absent, occluded, or too ambiguous to be reliably identified. 
To suppress such errors, we reuse the scoring function in Eq.~\ref{eq:binary_verification} to determine the valid temporal extent of the propagated sequence. 
For each propagated mask $\boldsymbol{M}_t$, the frame-level validity indicator is:
\begin{equation}
    c_t = \mathbb{I}\left(\psi(\mathcal{I}_t, {\boldsymbol{M}}_t, \mathcal{T}) > \delta\right).
    \label{eq:prompt_validity}
\end{equation}
The expansion stops once an invalid frame is encountered in the corresponding direction. 
The left and right temporal boundaries are determined as:
\begin{equation}
    \hat{t}_s = \min \{t \leq t^{*} \mid c_j = 1, \forall j \in [t,t^{*}]\}.
    \label{eq:left_boundary}
\end{equation}
\begin{equation}
    \hat{t}_e = \max \{t \geq t^{*} \mid c_j = 1, \forall j \in [t^{*},t]\}.
    \label{eq:right_boundary}
\end{equation}
The final verified mask sequence is:
\begin{equation}
    \mathcal{M} = \{{\boldsymbol{M}}_t\}_{t=\hat{t}_s}^{\hat{t}_e}.
    \label{eq:final_prompt_result}
\end{equation}

The proposal-based and prompt-based strategies are complementary. 
The former benefits from global tube proposal generation, while the latter can be used independently and also provides a direct alternative when reliable tube proposals are unavailable. 
Both strategies preserve the training-free nature of RSVG-ZeroOV and enable zero-shot open-vocabulary grounding in remote sensing videos.

\section{Experiments}
\label{sec:experiments}
\subsection{Datasets}
To thoroughly evaluate the effectiveness and generalization ability of our method, we consider both remote sensing image grounding benchmarks and video grounding datasets.

\noindent \textbf{Remote Sensing Image Grounding.}
We evaluate our method on two widely used benchmarks for remote sensing visual grounding (RSVG), namely RRSIS-D \cite{liu2024rotated} and RISBench \cite{dong2024cross}. 
Each sample in these datasets consists of a remote sensing image, a referring expression, and the corresponding annotation, which can be either a bounding box or a segmentation mask. 
Both datasets are extended from the RSREC dataset, enabling comprehensive evaluation under two settings: referring expression-based object detection (RSREC) and referring expression-based segmentation (RSRES). 
These benchmarks cover diverse scene types and object categories, providing a solid testbed for evaluating open-vocabulary grounding performance in remote sensing imagery.

\noindent \textbf{Remote Sensing Video Grounding.}
To evaluate the spatio-temporal grounding capability of our framework in remote sensing scenarios, we conduct experiments on the UAV-SAVG dataset \cite{zhan2025does}, which is specifically designed for spatial aerial video grounding. 
This dataset contains low-altitude aerial videos with natural language queries, requiring models to localize target objects consistently across frames. It provides a realistic and challenging benchmark for video-based remote sensing grounding.

\noindent \textbf{General Video Grounding.}
Given the limited availability of large-scale remote sensing video grounding datasets, we further evaluate our method on three widely used general-domain video grounding benchmarks, including HC-STVGv1, HC-STVGv2\cite{tang2021human} and VidSTG\cite{zhang2020does}. 
These datasets contain diverse dynamic scenes and complex referring expressions, enabling us to assess the generalization capability of our method beyond remote sensing domains. 
This cross-domain evaluation further demonstrates the robustness and scalability of our training-free framework for spatio-temporal grounding.

\subsection{Experimental Settings}
\noindent \textbf{Evaluation Metrics.}
We adopt different evaluation metrics for image-level and video-level grounding.
For remote sensing image grounding, we follow prior works~\cite{li2024language, liu2024rotated} and report three widely used metrics, including mean Intersection-over-Union (mIoU), overall Intersection-over-Union (oIoU), and Precision@X (P@X) with $X \in \{0.3,0.5,0.7\}$. Specifically, \textbf{mIoU} computes the average IoU between the prediction and ground truth over all samples, reflecting the average grounding quality at the instance level. \textbf{oIoU} is computed as the ratio between the total intersection area and the total union area over the entire dataset, which emphasizes the overall localization performance at the dataset level. \textbf{P@X} measures the proportion of samples whose IoU exceeds a predefined threshold $X$, thereby evaluating the reliability of predictions under different precision requirements. Following standard practice, in the RSREC task, the predicted bounding box is obtained from the minimum bounding rectangle of the predicted RSRES mask.
For video grounding, following prior work~\cite{yang2022tubedetr, zhan2025does}, we define the video Intersection-over-Union (vIoU) as $\mathrm{vIoU} = 1/N_f\sum_{t=1}^{N_f} \mathrm{IoU}(\hat{b}_t, b_t)$, where $N_f$ denotes the total number of frames in a video, and $\hat{b}_t$ and $b_t$ represent the predicted and ground-truth bounding boxes at frame $t$, respectively. Based on vIoU, we report two metrics for evaluating overall video grounding performance: \textbf{m\_vIoU} and \textbf{vIoU@R}. Here, \textbf{m\_vIoU} denotes the average vIoU over all video samples, while \textbf{vIoU@R} measures the proportion of videos whose vIoU exceeds a threshold $R$. These metrics primarily assess the global spatio-temporal localization accuracy of a model over an entire video. To further evaluate frame-level localization stability, we additionally adopt frame accuracy (\textbf{fAcc}). A frame is considered correctly localized if its frame-level IoU is greater than 0.5. Based on this criterion, \textbf{fAcc} is defined as the proportion of correctly localized frames in a video. We further report \textbf{m\_fAcc}, which averages fAcc over all videos, and \textbf{fAcc@R}, which measures the proportion of videos whose fAcc exceeds a threshold $R$. Following common practice, we set $R$ to 0.3 and 0.5 in our experiments.

\begin{table*}[!t]
\centering
\caption{Comparisons with SOTA methods on \textbf{RRSIS-D} test set. 
‘\textit{W-S}’ and ‘\textit{Z-S}’ denote the weakly-supervised and zero-shot settings, respectively.
‘\textit{w/ VLM}’ indicates that the cross-attention from the VLM is used to replace the original cross-attention.
‘\textit{w/ Refine}’ indicates that post-processing operations are applied to further refine the segmentation results.}
\resizebox{\linewidth}{!}{
\begin{tabular}{c|l|ccccc|ccccc}
    \specialrule{.1em}{.3em}{.3em} 
    \multicolumn{2}{c|}{}      & \multicolumn{5}{c|}{RSREC}    & \multicolumn{5}{c}{RSRES}\\
    \cmidrule(r){3-12}       
    \multicolumn{2}{l|}{Method}                        & Pr@0.3& Pr@0.5& Pr@0.7& mIoU  & oIoU  &Pr@0.3 & Pr@0.5 & Pr@0.7& mIoU  & oIoU\\
    \specialrule{.1em}{.3em}{.3em} 
    \multirow{3}{*}{\textit{W-S}} 
    & TRIS         (ICCV'23)                           & 14.62 & 3.79  & 0.37  & 13.20 & 13.14 & 15.02 & 4.91   & 1.26  & 13.11 & 15.40   \\ 
    & SAG          (ICCV'23)                           & 9.39  & 2.10  & 0.29  & 9.22  & 9.21  & 11.63 & 4.02   & 0.60  & 11.10 & 11.13    \\ 
    & QueryMatch   (MM'24)                             & 22.04 & 16.22 & 12.10 & 17.21 & 15.26 & 20.97 & 15.54  & 10.62 & 15.73 & 10.73    \\ 
    \midrule 
    \multirow{7}{*}{\textit{Z-S}}  
    & VLM (Baseline)                                   & 43.35 & 28.66 & 14.27 & 30.90 & 29.53 & -     & -      & -     & -     & -      \\ 
    & DiﬀSegmenter (TIP'25)                            & 8.22  & 1.69  & 0.20  & 8.70  & 8.63  & 2.18  & 0.23   & 0.00  & 4.47  & 4.35    \\ 
    & \textcolor{gray}{DiﬀSegmenter (\textit{w/ VLM})} & \textcolor{gray}{39.80} & \textcolor{gray}{24.29} & \textcolor{gray}{13.01} & \textcolor{gray}{28.47} & \textcolor{gray}{26.92} & \textcolor{gray}{24.99} & \textcolor{gray}{9.48} & \textcolor{gray}{1.61} & \textcolor{gray}{19.03} & \textcolor{gray}{15.19}\\
    & DiffPNG      (ECCV'24)                           & 18.93 & 8.45  & 3.30  & 14.84 & 12.54 & 17.70 & 8.27   & 3.02  & 14.86 & 14.71   \\ 
    & \textcolor{gray}{DiffPNG (\textit{w/ VLM})}      & \textcolor{gray}{37.92} & \textcolor{gray}{23.18} & \textcolor{gray}{11.35} & \textcolor{gray}{26.93} & \textcolor{gray}{25.65} & \textcolor{gray}{19.25} & \textcolor{gray}{6.78}   & \textcolor{gray}{0.95}  & \textcolor{gray}{15.49} & \textcolor{gray}{10.55}   \\
    & OV-VG (arXiv'24)                                 & 27.78 & 19.16 & 12.96 & 22.07 & 18.34 & -     & -      & -     & -     & -      \\
    & GroundVLP (AAAI'24)                              & 21.66 & 16.83 & 12.27 & 17.14 & 17.51 & -     & -      & -     & -     & -      \\
    & \textbf{RSVG-ZeroOV (Ours)}                      & \textbf{45.71} & 30.15 & 16.74 & \underline{32.92} & \textbf{32.94} & 30.97 & 12.84  & 2.36  & 21.85 & 18.85   \\  
    \midrule 
    \multirow{6}{*}{\textit{Z-S}}  
    & VLM (Baseline)                                   & 43.60 & \underline{30.35} & \underline{16.76} & 31.93 & 28.38 & \underline{38.18} & \underline{24.68}  & \underline{11.00} & \underline{25.72} & \underline{20.46}   \\ 
    & DiﬀSegmenter (TIP'25)                            & 8.16  & 1.55  & 0.17  & 8.60  & 8.58  & 3.02  & 0.32   & 0.03  & 4.60  & 4.70\\ 
    & \textcolor{gray}{DiﬀSegmenter (\textit{w/ VLM})} & \textcolor{gray}{38.41} & \textcolor{gray}{25.11} & \textcolor{gray}{13.96} & \textcolor{gray}{28.50} & \textcolor{gray}{24.57} & \textcolor{gray}{32.92} & \textcolor{gray}{19.42} & \textcolor{gray}{9.14} & \textcolor{gray}{23.73} & \textcolor{gray}{18.00}\\
    \multirow{4}{*}{\textit{w/ Refine}}  
    & DiffPNG      (ECCV'24)          & 19.59 & 10.31  & 5.03  & 15.71 & 13.11 & 19.65 & 11.32 & 6.23  & 16.22 & 13.77   \\ 
    & \textcolor{gray}{DiffPNG (\textit{w/ VLM})}      & \textcolor{gray}{32.43}    & \textcolor{gray}{21.29}  & \textcolor{gray}{13.04} & \textcolor{gray}{24.89} & \textcolor{gray}{20.79} & \textcolor{gray}{27.03} & \textcolor{gray}{17.64}  & \textcolor{gray}{\underline{11.00}} & \textcolor{gray}{20.99} & \textcolor{gray}{13.53}   \\
    & OV-VG (arXiv'24)                                 & 27.23 & 16.20 & 13.27 & 21.62 & 17.18 & 20.68 & 15.51 & 9.05  & 16.17 & 9.71  \\
    & GroundVLP (AAAI'24)                              & 21.26 & 16.20 & 11.32 & 16.51 & 16.88 & 18.07 & 13.33 & 7.24  & 13.14 & 11.23 \\
    & \textbf{RSVG-ZeroOV (Ours)}                      & \underline{45.70} & \textbf{31.39}  & \textbf{17.63} & \textbf{34.49} & \underline{31.28} & \textbf{40.01} & \textbf{27.39}  & \textbf{13.38} & \textbf{28.35} & \textbf{22.83}  \\ 
    \specialrule{.1em}{.3em}{.3em} 
\end{tabular}}
\label{tab:RRSIS-D}
\end{table*}

\noindent \textbf{Comparison with State-of-the-Art Methods.}
We compare RSVG-ZeroOV with a diverse set of state-of-the-art methods for both image and video grounding. 
For image grounding, we include representative weakly-supervised and zero-shot methods, including TRIS~\cite{liu2023referring}, SAG~\cite{hong2023improving}, QueryMatch~\cite{chen2024querymatch}, DiffSegmenter~\cite{wang2025diffusion}, DiffPNG~\cite{yang2024exploring}, OV-VG~\cite{wang2024ov}, and GroundVLP~\cite{shen2024groundvlp}. 
For a fair comparison, we further implement two adapted variants of DiffSegmenter and DiffPNG by replacing their original cross-attention maps with those extracted from our adopted VLM and applying the same refinement strategy as in our framework. 
In addition, we compare with several strong generic pixel-level vision-language models, including GeoChat~\cite{kuckreja2024geochat}, Qwen2.5-VL~\cite{Qwen2.5-VL}, LISA~\cite{lai2024lisa}, and NExT-Chat~\cite{zhang2023next}, to evaluate the grounding capability of generic multimodal models in remote sensing scenarios.
For video grounding, we conduct comprehensive comparisons with state-of-the-art methods from both general-domain and remote sensing scenarios. 
Specifically, we compare with fully supervised spatio-temporal video grounding methods, including Co-grounding~\cite{song2021co}, DCNet~\cite{cao2022correspondence}, TubeDETR~\cite{yang2022tubedetr}, STCAT~\cite{jin2022embracing}, SGFDN~\cite{wang2023efficient}, CG-STVG~\cite{gu2024context}, VideoGrounding-DINO~\cite{wasim2024videogrounding}, and the recent remote sensing video grounding method SAVG-DETR~\cite{zhan2025does}. 
We also include weakly-supervised methods, including WINNER~\cite{li2023winner}, VEM~\cite{jin2024weakly}, CoPAL~\cite{kumar2025contextual}, and STPro~\cite{garg2025stpro}, as well as zero-shot methods, including ReCLIP~\cite{subramanian2022reclip}, RedCircle~\cite{shtedritski2023does}, and E3M~\cite{bao2024e3m}. 
These comparisons cover fully supervised, weakly-supervised, and zero-shot settings, enabling a comprehensive evaluation of the proposed Video RSVG-ZeroOV across different grounding paradigms.

\noindent \textbf{Implementation Details.}
Our RSVG-ZeroOV offers a zero-shot solution, requiring only inference without the need for any training images or annotations.
We employ the pre-trained Qwen2.5-VL~\cite{Qwen2.5-VL} and Stable Diffusion V1.4~\cite{rombach2022high} as our VLM and DM, respectively.
For Stable Diffusion, we use a guidance scale of 7.5 with 1,000 total diffusion steps, and perform DDIM-based sampling with 20 steps.
The DDIM noise schedule follows a scaled linear progression from 0.00085 to 0.012.
We set $K=7$ for seed selection, $\tau=0.3$ for response thresholding, $\alpha=0.4$ for binarization, and $\delta=0.05$ for VLM-based temporal verification.
All experiments are implemented with PyTorch on a DELL EMC DSS8440 server equipped with a Xeon Silver 4210R CPU, 128 GB RAM, and four NVIDIA GeForce RTX 4090 GPUs with 24 GB of VRAM.

\begin{table}[!t]
\centering
\caption{Comparisons with SOTA generic pixel-level VLMs and a remote sensing VLM on \textbf{RRSIS-D} test set.}
\resizebox{\linewidth}{!}{
\begin{tabular}{l|cc|cc}
    \specialrule{.1em}{.3em}{.3em} 
    \multicolumn{1}{c|}{}                         & \multicolumn{2}{c|}{RSREC} & \multicolumn{2}{c}{RSRES}\\
    \cmidrule(r){2-5}       
    \multicolumn{1}{l|}{Method} & Pr@0.5 & mIoU & Pr@0.5 & mIoU\\
    \specialrule{.1em}{.3em}{.3em} 
    GeoChat     (CVPR'24) + SAM          & 27.61 & 32.53 & 18.85 & 24.92 \\
    Qwen2.5-VL   (arXiv'25) + SAM        & 30.35 & 31.93 & 24.68 & 20.46 \\
    LISA        (CVPR'24)                & 25.80 & 27.78 & 24.51 & 26.78 \\
    NExT-Chat   (ICML'24)                & -     & -     & 26.37 & 24.98 \\
    \textbf{RSVG-ZeroOV (Ours)}          & \textbf{31.39} & \textbf{34.49} & \textbf{27.39} & \textbf{28.35} \\ 
    \specialrule{.1em}{.3em}{.3em} 
\end{tabular}}
\label{tab:appendixresults}
\end{table}

\subsection{Comparison With Others.}
\subsubsection{Performance on Remote Sensing Image Grounding}

\noindent \textbf{Comparison Results on RRSIS-D.}
As shown in Tab.~\ref{tab:RRSIS-D}, RSVG-ZeroOV (\textit{w/ Refine}) achieves SOTA performance under the zero-shot setting, surpassing all weakly-supervised and zero-shot methods.
Compared with the strongest weakly-supervised baseline QueryMatch, RSVG-ZeroOV improves mIoU by 17.28 points on RSREC and 12.62 points on RSRES, despite using no task-specific training data. 
Compared with other zero-shot approaches, \textit{i.e.}, DiffSegmenter (\textit{w/ VLM}) and DiffPNG (\textit{w/ VLM}), our method also improves mIoU by 5.99--9.60 points on RSREC and 4.62--7.36 points on RSRES. 
This highlights the effectiveness of our framework and its related components beyond VLM alone.
Furthermore, we compare our method against several SOTA generic VLMs and a remote sensing VLM, as shown in Table~\ref{tab:appendixresults}. RSVG-ZeroOV achieves the best performance across all metrics on both RSREC and RSRES, demonstrating superior grounding ability even compared to models tailored for pixel-level reasoning or remote sensing tasks.

\begin{table*}[!t]
\centering
\caption{Comparisons with SOTA methods on \textbf{RISBench} test set. 
‘\textit{W-S}’ and ‘\textit{Z-S}’ denote the weakly-supervised and zero-shot settings, respectively.
‘\textit{w/ VLM}’ indicates that the cross-attention from the VLM is used to replace the original cross-attention.
‘\textit{w/ Refine}’ indicates that post-processing operations are applied to further refine the segmentation results.}
\resizebox{\linewidth}{!}{
\begin{tabular}{c|l|ccccc|ccccc}
    \specialrule{.1em}{.3em}{.3em} 
    \multicolumn{2}{c|}{}      & \multicolumn{5}{c|}{RSREC}    & \multicolumn{5}{c}{RSRES}\\
    \cmidrule(r){3-12}       
    \multicolumn{2}{l|}{Method}                             & Pr@0.3& Pr@0.5 & Pr@0.7& mIoU  & oIoU  &Pr@0.3 & Pr@0.5& Pr@0.7& mIoU  & oIoU\\
    \specialrule{.1em}{.3em}{.3em} 
    \multirow{3}{*}{\textit{W-S}} 
    & TRIS         (ICCV'23)          & 19.27 & 8.89  & 2.86  & 14.63 & 15.74 & 14.23 & 4.40   & 0.92  & 11.46 & 16.06   \\ 
    & SAG          (ICCV'23)          & 5.83  & 1.78  & 0.33  & 7.12  & 7.10  & 8.79  & 2.55   & 0.35  & 9.31  & 9.36    \\ 
    & QueryMatch   (MM'24)            & 31.79 & 27.74 & 23.32 & 26.72 & 15.86 & 31.06 & 26.68  & \textbf{20.27} & 24.59 & 10.77    \\ 
    \midrule 
    \multirow{7}{*}{\textit{Z-S}}  
    & VLM (Baseline)                  & 48.62 & 35.08 & 20.26 & 34.60 & 30.15 & -     & -      & -     & -     & -      \\ 
    & DiﬀSegmenter (TIP'25)           & 6.03  & 1.87  & 0.40  & 7.14  & 7.00  & 2.44  & 0.46  & 0.06  & 4.00  & 3.74    \\ 
    & \textcolor{gray}{DiﬀSegmenter (\textit{w/ VLM})}& \textcolor{gray}{27.17} & \textcolor{gray}{15.93} & \textcolor{gray}{7.98} & \textcolor{gray}{21.61} & \textcolor{gray}{19.03} & \textcolor{gray}{28.65} & \textcolor{gray}{11.04} & \textcolor{gray}{2.50} & \textcolor{gray}{21.37} & \textcolor{gray}{\underline{21.87}} \\
    & DiffPNG      (ECCV'24)          & 12.14 & 6.03  & 2.72  & 11.10 & 9.01  & 13.44 & 5.79  & 1.93  & 11.85 & 11.79   \\ 
    & \textcolor{gray}{DiffPNG (\textit{w/ VLM})}     & \textcolor{gray}{34.60} & \textcolor{gray}{21.54} & \textcolor{gray}{11.62} & \textcolor{gray}{26.18} & \textcolor{gray}{23.58} & \textcolor{gray}{23.50} & \textcolor{gray}{8.73} & \textcolor{gray}{1.76} & \textcolor{gray}{18.49} & \textcolor{gray}{14.45} \\
    & OV-VG (arXiv'24)                                & 26.83 & 22.16 & 18.31 & 22.65 & 15.64 & -     & -      & -     & -     & -      \\
    & GroundVLP (AAAI'24)                             & 23.13 & 19.90 & 17.15 & 19.76 & 16.46 & -     & -      & -     & -     & -      \\
    &  \textbf{RSVG-ZeroOV (Ours)}                    & \underline{50.17} & 35.08 & 19.03 & 36.20 & \textbf{34.37} & 32.71 & 14.74  & 3.73  & 22.69 & 21.75   \\  
    \midrule 
    \multirow{6}{*}{\textit{Z-S}}  
    & VLM (Baseline)                  & 49.07 & \underline{38.09} & \textbf{26.57} & \underline{37.74} & 30.68 & \underline{43.03} & \underline{30.38} & 18.04 & \underline{30.45} & 21.30  \\ 
    & DiﬀSegmenter (TIP'25)           & 6.05  & 1.99  & 0.45  & 7.11  & 6.96  & 2.65  & 0.73   & 0.17  & 3.88  & 3.79 \\ 
    & \textcolor{gray}{DiﬀSegmenter (\textit{w/ VLM})}& \textcolor{gray}{26.74} & \textcolor{gray}{16.74} & \textcolor{gray}{9.47} & \textcolor{gray}{22.01} & \textcolor{gray}{17.91} & \textcolor{gray}{21.39} & \textcolor{gray}{12.02} & \textcolor{gray}{6.40} & \textcolor{gray}{17.65} & \textcolor{gray}{12.87} \\
    \multirow{4}{*}{\textit{w/ Refine}}  
    & DiffPNG      (ECCV'24)          & 9.34  & 3.76  & 1.38  & 8.43  & 8.66  & 3.99  & 1.73  & 0.76  & 4.53 & 5.81   \\ 
    & \textcolor{gray}{DiffPNG (\textit{w/ VLM})}     & \textcolor{gray}{22.07}    & \textcolor{gray}{12.19}  & \textcolor{gray}{6.00} & \textcolor{gray}{17.16} & \textcolor{gray}{17.40} & \textcolor{gray}{14.74} & \textcolor{gray}{5.65}  & \textcolor{gray}{1.60} & \textcolor{gray}{12.19} & \textcolor{gray}{8.44}   \\
    & OV-VG (arXiv'24)                                & 26.90 & 22.40 & 18.50 & 22.85 & 15.50 & 23.67 & 17.75 & 12.25 & 18.16 & 9.20  \\
    & GroundVLP (AAAI'24)                             & 23.13 & 19.91 & 16.37 & 19.19 & 16.17 & 20.78 & 15.82 & 10.95 & 15.58 & 10.08 \\
    &  \textbf{RSVG-ZeroOV (Ours)}      & \textbf{50.77} & \textbf{38.90}  & \underline{24.93} & \textbf{38.87} & \underline{34.30} & \textbf{44.30} & \textbf{31.03}  & \underline{18.61} & \textbf{31.84} & \textbf{26.35}  \\ 
    \specialrule{.1em}{.3em}{.3em} 
\end{tabular}}
\label{tab:RISBench}
\end{table*}

\noindent \textbf{Comparison Results on RISBench.} 
We further evaluate RSVG-ZeroOV on RISBench, which contains longer and more semantically complex referring expressions than RRSIS-D. 
As shown in Table~\ref{tab:RISBench}, RSVG-ZeroOV maintains strong zero-shot performance and achieves the best results on most RSREC and RSRES metrics after refinement. 
Compared with the strongest weakly-supervised baseline QueryMatch, our method improves RSREC mIoU by 12.15 points and RSRES mIoU by 7.25 points. 
It also substantially improves oIoU over zero-shot grounding baselines. 
Compared with OV-VG and GroundVLP, RSVG-ZeroOV improves RSREC oIoU by 18.80 and 18.13 points, respectively, and improves RSRES oIoU by 17.15 and 16.27 points, respectively. 
These results show that RSVG-ZeroOV remains effective when referring expressions involve more complex semantic constraints. 
The improvements on oIoU are particularly important, as this metric reflects whether the predicted mask can preserve complete object regions rather than only covering highly discriminative parts. 
This property is also beneficial for the subsequent video extension, where a complete and query-consistent key-frame mask provides a more reliable spatial anchor for temporal propagation.

\begin{figure*}[!t]
    \centering
    \includegraphics[width=\linewidth]{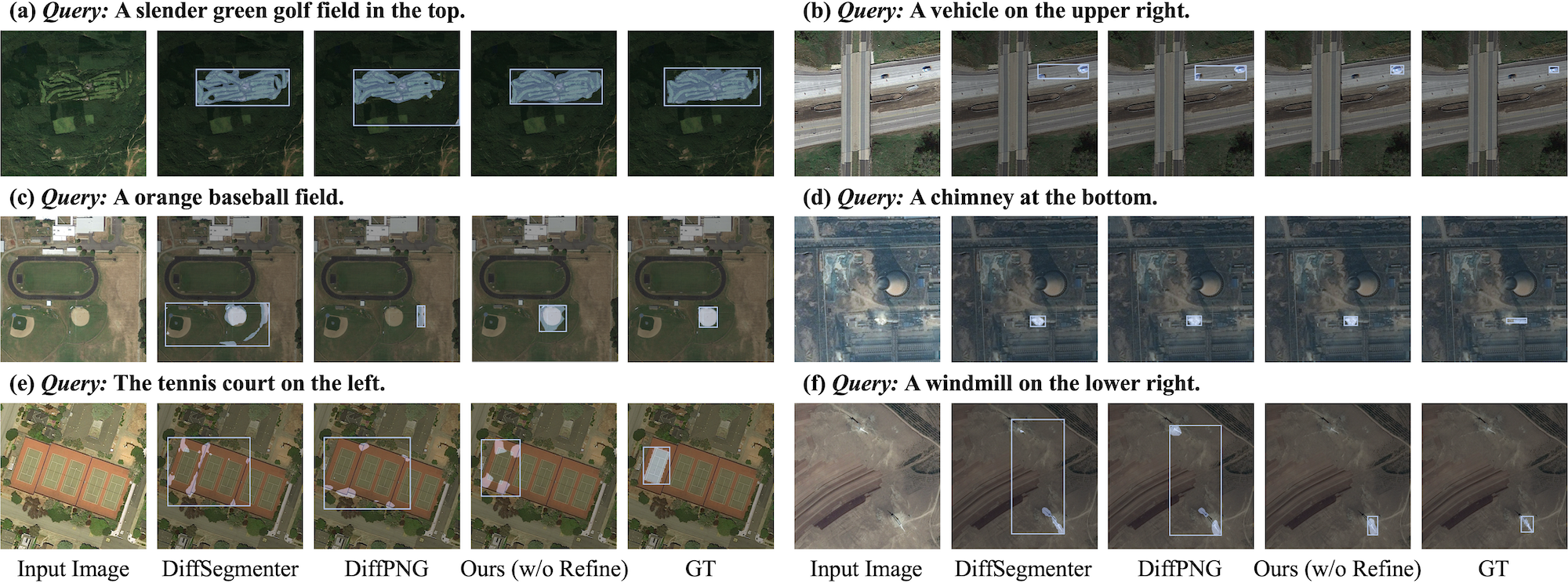}
    \caption{Qualitative results of different methods for image-based RSVG.}
    \label{fig:visualization_image}
\end{figure*}

\begin{table*}[!t]
\centering
\caption{Comparisons with SOTA methods on \textbf{UAV-SAVG} test set. ‘\textit{F-S}’ and ‘\textit{Z-S}’ denote the fully-supervised and zero-shot settings, respectively. ‘\textit{w/ Strategy 1}’ and ‘\textit{w/ Strategy 2}’ indicate the two proposed temporal strategies for video grounding, respectively.}
\setlength{\tabcolsep}{2.5mm}
\resizebox{\linewidth}{!}{
\begin{tabular}{c|l|cccccc}
\specialrule{.1em}{.3em}{.3em}
\multicolumn{2}{l|}{Method} & m\_vIoU & vIoU@0.3 & vIoU@0.5 & m\_fAcc & fAcc@0.3 & fAcc@0.5 \\
\midrule
\multirow{8}{*}{\textit{F-S}}
& Co-grounding (CVPR'21)                    & 10.24 & 21.66 & 6.11 & 11.17 & 16.29 & 8.40 \\
& DCNet (ACMMM'22)                          & 11.65 & 23.58 & 8.79 & 13.10 & 17.64 & 9.21 \\
& TubeDETR (CVPR'22)                        & 22.60 & 32.91 & 20.49 & 23.84 & 29.69 & 22.00 \\
& STCAT (NeurIPS'22)                        & 24.14 & 35.51 & 22.48 & 27.17 & 33.39 & 25.36 \\
& SGFDN (ACMMM'23)                          & 20.13 & 28.16 & 15.47 & 19.13 & 22.71 & 17.39 \\
& CG-STVG (CVPR'24)                         & 21.23 & 28.82 & 19.04 & 22.32 & 26.24 & 20.41 \\
& VideoGrounding-DINO (CVPR'24)             & 23.83 & 33.84 & 19.92 & 25.80 & 31.72 & 23.00 \\
& SAVG-DETR (NeurIPS'25)                    & \underline{27.15} & \textbf{38.18} & \underline{22.85} & \underline{28.82} & \underline{35.85} & \underline{26.55} \\
\midrule 
\multirow{8}{*}{\textit{Z-S}}
& ReCLIP (ACL'22)                           & 4.52 & 0.00 & 0.00 & -- & -- & -- \\
& RedCircle (CVPR'23)                       & 5.31 & 0.00 & 0.00 & -- & -- & -- \\
\cmidrule(l){2-8}
& \multicolumn{5}{l}{\textbf{\textit{w/ Strategy 1}:}} \\
& Qwen-2.5-VL (arXiv'25)                    & 17.71 & 24.50 & 17.65 & 21.14 & 25.08 & 22.47 \\
& \textbf{Video RSVG-ZeroOV (Ours)}         
& \textbf{27.19} {\scriptsize\textcolor{darkgreen}{$\uparrow$9.48}} 
& 35.62 {\scriptsize\textcolor{darkgreen}{$\uparrow$11.12}} 
& \textbf{29.28} {\scriptsize\textcolor{darkgreen}{$\uparrow$11.63}} 
& \textbf{33.77} {\scriptsize\textcolor{darkgreen}{$\uparrow$12.63}} 
& \textbf{37.87} {\scriptsize\textcolor{darkgreen}{$\uparrow$12.79}} 
& \textbf{34.61} {\scriptsize\textcolor{darkgreen}{$\uparrow$12.14}} \\
\cmidrule(l){2-8}
& \multicolumn{5}{l}{\textbf{\textit{w/ Strategy 2}:}} \\
& Qwen-2.5-VL (arXiv'25)                    & 13.14 & 18.31 & 14.71 & 16.43 & 18.66 & 17.21 \\
& \textbf{Video RSVG-ZeroOV (Ours)}         
& 21.96 {\scriptsize\textcolor{darkgreen}{$\uparrow$8.82}} 
& 27.91 {\scriptsize\textcolor{darkgreen}{$\uparrow$9.60}} 
& 20.80 {\scriptsize\textcolor{darkgreen}{$\uparrow$6.09}} 
& 24.57 {\scriptsize\textcolor{darkgreen}{$\uparrow$8.14}} 
& 28.78 {\scriptsize\textcolor{darkgreen}{$\uparrow$10.12}} 
& 24.84 {\scriptsize\textcolor{darkgreen}{$\uparrow$7.63}} \\
\specialrule{.1em}{.3em}{.3em}
\end{tabular}}
\label{tab:UAV-SAVG}
\end{table*}

\begin{figure*}[!t]
    \centering
    \includegraphics[width=\linewidth]{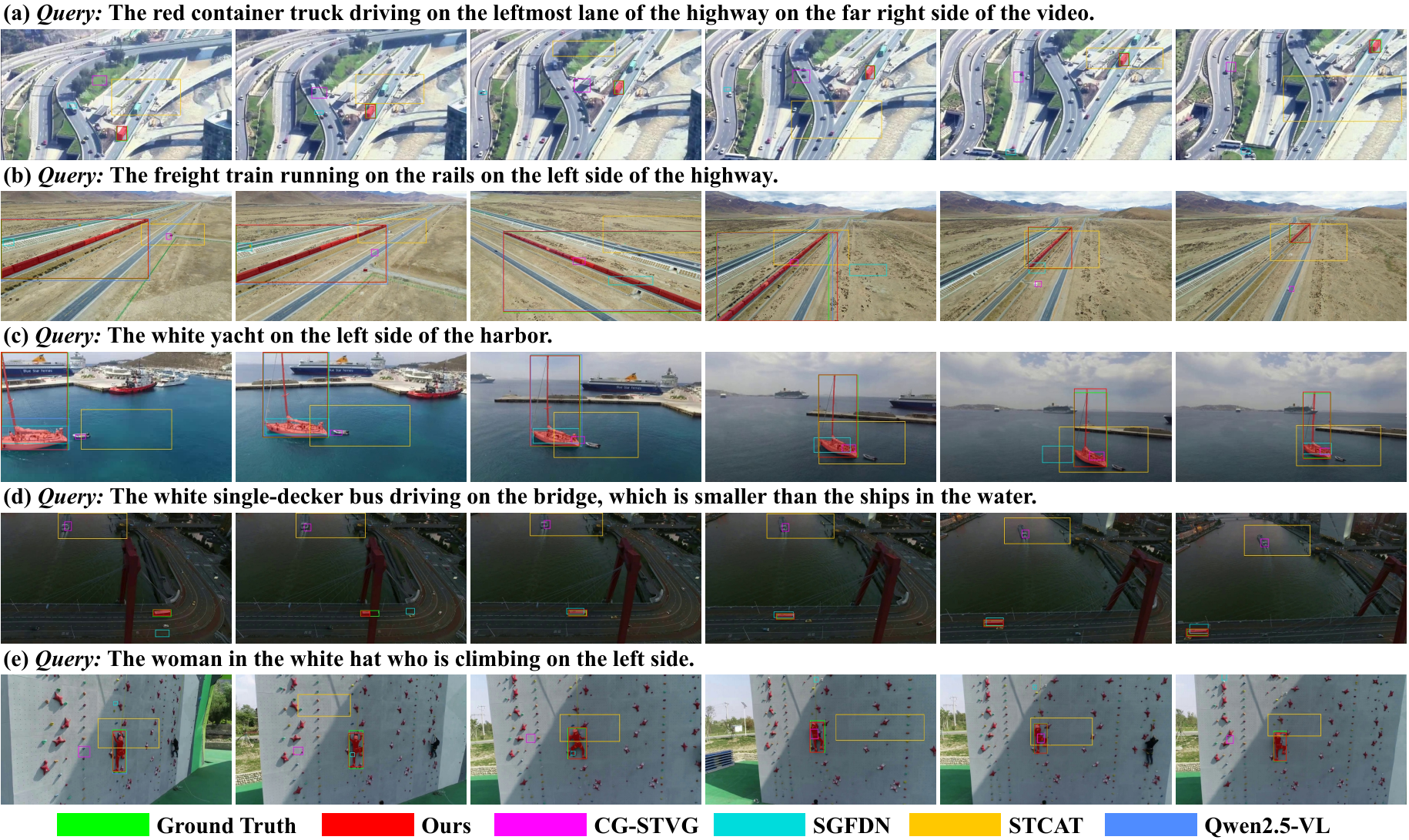}
    \caption{Qualitative results of different methods for video-based RSVG.}
    \label{fig:UAV-SAVG}
\end{figure*}

\noindent \textbf{Visualization Results.}
We present qualitative results of the proposed RSVG-ZeroOV in Figure \ref{fig:visualization_image}, comparing with representative zero-shot baselines, including DiffSegmenter and DiffPNG. 
Our method consistently produces more accurate and complete grounding results across diverse remote sensing scenes. 
In contrast, existing methods often suffer from incomplete localization or semantic ambiguity. 
For example, in Figure \ref{fig:visualization_image} \textcolor{red}{(a)} and \textcolor{red}{(c)}, DiffSegmenter and DiffPNG tend to focus only on partial regions of the target objects or exhibit fragmented predictions, while our method successfully captures the full spatial extent of the referred objects, demonstrating stronger structural awareness.
Moreover, our approach shows improved robustness in complex scenarios involving small objects and cluttered backgrounds. 
In Figure \ref{fig:visualization_image} \textcolor{red}{(b)} and \textcolor{red}{(d)}, baseline methods either miss the target or produce scattered activations, whereas our method can precisely localize the referred objects with clearer boundaries. 
This indicates the effectiveness of our attention refinement strategy in suppressing irrelevant regions.
In addition, our method demonstrates superior capability in understanding spatial relations described in natural language. 
As illustrated in Figure \ref{fig:visualization_image} \textcolor{red}{(e)} and \textcolor{red}{(f)}, our model accurately identifies targets specified by relative positions (e.g., “on the left” or “on the lower right”), while competing methods often fail to align such spatial cues with visual regions.
Overall, these visualizations validate that RSVG-ZeroOV not only improves localization accuracy but also enhances structural completeness and semantic consistency, highlighting its effectiveness for open-vocabulary remote sensing visual grounding.

\subsubsection{Performance on Remote Sensing Video Grounding}
\noindent \textbf{Comparison Results on UAV-SAVG.}
Table~\ref{tab:UAV-SAVG} reports the comparison results on the UAV-SAVG test set. 
All variants of Video RSVG-ZeroOV are evaluated under the zero-shot setting, without using task-specific video grounding annotations. 
Despite this challenging setting, Video RSVG-ZeroOV with Strategy 1 achieves the best performance on five out of six metrics among all compared methods, including fully-supervised approaches. 
Compared with the strongest fully-supervised baseline SAVG-DETR, our method obtains comparable m\_vIoU and clearly improves several stricter and frame-sensitive metrics, including vIoU@0.5, m\_fAcc, fAcc@0.3, and fAcc@0.5, with gains of 6.43, 4.95, 2.02, and 8.06 points, respectively. 
This indicates that the proposed training-free framework can produce accurate frame-wise localization of the referred object, rather than only recovering coarse spatio-temporal regions.

\begin{table*}[!t]
\centering
\caption{Comparisons with SOTA methods on the \textbf{HC-STVGv1} and \textbf{HC-STVGv2} test sets. ‘\textit{F-S}’, ‘\textit{W-S}’ and ‘\textit{Z-S}’ denote the fully-supervised, weakly-supervised and zero-shot settings, respectively. ‘\textit{w/ Strategy 1}’ and ‘\textit{w/ Strategy 2}’ indicate the two proposed temporal strategies for video grounding, respectively.}
\setlength{\tabcolsep}{2mm}
\resizebox{\linewidth}{!}{
\begin{tabular}{c|l|ccc|ccc}
\specialrule{.1em}{.3em}{.3em} 
\multicolumn{2}{c|}{} & \multicolumn{3}{c|}{HCSTVG-v1} & \multicolumn{3}{c}{HCSTVG-v2} \\
\cmidrule(r){3-5} \cmidrule(r){6-8}
\multicolumn{2}{l|}{Method} & m\_vIoU & vIoU@0.3 & vIoU@0.5 & m\_vIoU & vIoU@0.3 & vIoU@0.5 \\
\specialrule{.1em}{.3em}{.3em} 
\multirow{5}{*}{\textit{F-S}} 
& TubeDETR (CVPR'22)            & \textcolor{gray}{32.40} & \textcolor{gray}{49.80} & \textcolor{gray}{23.50} & \textcolor{gray}{36.40} & \textcolor{gray}{58.80} & \textcolor{gray}{30.60} \\
& STCAT (NeurIPS'22)            & \textcolor{gray}{35.09} & \textcolor{gray}{57.67} & \textcolor{gray}{30.09} & - & - & - \\
& CSDVL (CVPR'23)               & \textcolor{gray}{36.90} & \textcolor{gray}{62.20} & \textcolor{gray}{34.80} & \textcolor{gray}{38.70} & \textcolor{gray}{65.50} & \textcolor{gray}{33.80} \\
& CG-STVG (CVPR'24)             & \textcolor{gray}{38.40} & \textcolor{gray}{61.50} & \textcolor{gray}{36.30} & \textcolor{gray}{39.50} & \textcolor{gray}{64.50} & \textcolor{gray}{36.30} \\
& VideoGrounding-DINO (CVPR'24) & \textcolor{gray}{38.25} & \textcolor{gray}{62.47} & \textcolor{gray}{36.14} & \textcolor{gray}{39.88} & \textcolor{gray}{67.13} & \textcolor{gray}{34.49} \\
\midrule
\multirow{4}{*}{\textit{W-S}} 
& WINNER (CVPR'23) & 14.20 & 17.24 & 6.12  & -- & -- & -- \\
& VEM (ECCV'24)    & 14.64 & 18.60 & 5.75  & -- & -- & -- \\
& CosPaL (ICLR'25) & 22.10 & 31.80 & \textbf{19.60} & 22.20 & 31.40 & \textbf{18.90} \\
& STPro (CVPR'25)  & 17.56 & 26.98 & 12.93 & 19.99 & 31.70 & 14.55 \\
\midrule 
\multirow{10}{*}{\textit{Z-S}}
& ReCLIP (ACL'22)     & 14.36 & 18.28 & 4.91  & -- & -- & -- \\
& RedCircle (CVPR'23) & 9.15  & 7.76  & 1.55  & -- & -- & -- \\
& E3M (ECCV'24)       & 19.11 & 29.40 & 10.60 & -- & -- & -- \\
\cmidrule{2-8}
& \multicolumn{7}{l}{\textbf{\textit{w/ Strategy 1}:}} \\
& Qwen-2.5-VL (arXiv'25)            & 11.42 & 21.83 & 8.76 & 14.91 & 23.44 & 11.35 \\
& \textbf{Video RSVG-ZeroOV (Ours)} 
& \textbf{22.67} {\scriptsize\textcolor{darkgreen}{$\uparrow$11.25}} 
& \textbf{34.41} {\scriptsize\textcolor{darkgreen}{$\uparrow$12.58}} 
& \underline{16.58} {\scriptsize\textcolor{darkgreen}{$\uparrow$7.82}} 
& \textbf{25.12} {\scriptsize\textcolor{darkgreen}{$\uparrow$10.21}} 
& \textbf{36.26} {\scriptsize\textcolor{darkgreen}{$\uparrow$12.82}} 
& \underline{18.64} {\scriptsize\textcolor{darkgreen}{$\uparrow$7.29}} \\
\cmidrule{2-8}
& \multicolumn{7}{l}{\textbf{\textit{w/ Strategy 2}:}} \\
& Qwen-2.5-VL (arXiv'25)            & 10.86 & 15.82 & 7.92 & 12.13 & 17.20 & 9.47 \\
& \textbf{Video RSVG-ZeroOV (Ours)} 
& 18.34 {\scriptsize\textcolor{darkgreen}{$\uparrow$7.48}} 
& 26.09 {\scriptsize\textcolor{darkgreen}{$\uparrow$10.27}} 
& 12.84 {\scriptsize\textcolor{darkgreen}{$\uparrow$4.92}} 
& 23.45 {\scriptsize\textcolor{darkgreen}{$\uparrow$11.32}} 
& 27.41 {\scriptsize\textcolor{darkgreen}{$\uparrow$10.21}} 
& 13.28 {\scriptsize\textcolor{darkgreen}{$\uparrow$3.81}} \\
\specialrule{.1em}{.3em}{.3em}
\end{tabular}}
\label{tab:HC_STVG}
\end{table*}

Compared with zero-shot baselines, Video RSVG-ZeroOV also shows consistent advantages under both temporal strategies. 
With Strategy 1, our method improves over Qwen-2.5-VL by 9.48 points in m\_vIoU, 11.63 points in vIoU@0.5, 12.63 points in m\_fAcc, and 12.14 points in fAcc@0.5. 
Strategy 2 also brings clear improvements over its corresponding Qwen-2.5-VL baseline, demonstrating that the key-frame grounding result produced by RSVG-ZeroOV provides an effective spatial initialization for video-level propagation.

\noindent \textbf{Visualization Results.}
Figure~\ref{fig:UAV-SAVG} presents qualitative comparisons under challenging UAV-view scenarios. 
Overall, Video RSVG-ZeroOV achieves more accurate localization than competing methods. 
In Figure~\ref{fig:UAV-SAVG} \textcolor{red}{(a)--(c)}, our method reliably localizes the referred truck, train, and yacht, whereas Qwen2.5-VL often fails to identify the target and fully-supervised methods such as CG-STVG, SGFDN, and STCAT produce inaccurate results. 
These examples suggest that generic VLMs alone are insufficient for precise grounding, while supervised video grounding models may still struggle with small targets, elongated objects, and complex aerial scenes. 
Figure~\ref{fig:UAV-SAVG} \textcolor{red}{(d)} further highlights the importance of relational reasoning: the ships in the water serve only as reference objects for identifying the smaller white single-decker bus, but CG-STVG and STCAT are distracted by these contextual references. 
In contrast, our method correctly localizes the target and can further produce pixel-wise masks, providing finer-grained grounding results than box-only localization. 
These qualitative results demonstrate the effectiveness of our query-aware key-frame grounding and temporal propagation design under complex expressions and challenging aerial-view conditions.

\begin{table}[!t]
\centering
\caption{Comparisons with SOTA methods on the \textbf{VidSTG} test set. ‘\textit{F-S}’, ‘\textit{W-S}’ and ‘\textit{Z-S}’ denote the fully-supervised, weakly-supervised and zero-shot settings, respectively. ‘\textit{w/ Strategy 1}’ and ‘\textit{w/ Strategy 2}’ indicate the two proposed temporal strategies for video grounding, respectively.}
\setlength{\tabcolsep}{0.5mm}
\resizebox{\linewidth}{!}{
\begin{tabular}{l|cccc}
\specialrule{.1em}{.3em}{.3em} 
\multicolumn{1}{c|}{} & \multicolumn{4}{c}{VidSTG} \\
\cmidrule(r){2-5}
Method & m\_tIoU & m\_vIoU & vIoU@0.3 & vIoU@0.5 \\
\specialrule{.1em}{.3em}{.3em} 
\multicolumn{5}{l}{\textbf{\textit{F-S}:}} \\
TubeDETR (CVPR'22)            & \textcolor{gray}{48.10} & \textcolor{gray}{30.40} & \textcolor{gray}{42.50} & \textcolor{gray}{28.20} \\
STCAT (NeurIPS'22)            & \textcolor{gray}{50.82} & \textcolor{gray}{33.14} & \textcolor{gray}{46.20} & \textcolor{gray}{32.58} \\
CSDVL (CVPR'23)               & -                       & \textcolor{gray}{33.70} & \textcolor{gray}{47.20} & \textcolor{gray}{32.80} \\
CG-STVG (CVPR'24)             & \textcolor{gray}{51.40} & \textcolor{gray}{34.00} & \textcolor{gray}{47.70} & \textcolor{gray}{33.10} \\
VideoG-DINO (CVPR'24)         & \textcolor{gray}{51.97} & \textcolor{gray}{34.67} & \textcolor{gray}{48.11} & \textcolor{gray}{33.96} \\
\midrule
\multicolumn{5}{l}{\textbf{\textit{W-S}:}} \\
WINNER (CVPR'23) & - & 11.61 & 14.12 & 7.40 \\
VEM (ECCV'24)    & - & 14.45 & 18.57 & 8.76 \\
CosPaL (ICLR'25) & - & 16.00 & 20.10 & 13.10 \\
STPro (CVPR'25)  & - & 15.52 & 19.39 & \underline{12.69} \\
\midrule 
\multicolumn{5}{l}{\textbf{\textit{Z-S}:}} \\
ReCLIP (ACL'22)     & - & 14.21 & 17.54 & 7.86 \\
RedCircle (CVPR'23) & - & 8.56  & 7.61  & 0.93 \\
E3M (ECCV'24)       & - & \underline{16.21} & \underline{20.47} & 11.91 \\
\midrule
\multicolumn{5}{l}{\textbf{\textit{Z-S w/ Strategy 1}:}} \\
Qwen-2.5-VL (arXiv'25)            & 12.43 & 8.79  & 10.20 & 7.67 \\
\multirow{2}{*}{\textbf{Video RSVG-ZeroOV (Ours)}} 
& \textbf{30.16} & \textbf{17.71} & \textbf{21.12} & \textbf{15.26} \\
& {\scriptsize\textcolor{darkgreen}{$\uparrow$17.73}} 
& {\scriptsize\textcolor{darkgreen}{$\uparrow$8.92}} 
& {\scriptsize\textcolor{darkgreen}{$\uparrow$10.92}} 
& {\scriptsize\textcolor{darkgreen}{$\uparrow$7.59}} \\
\midrule
\multicolumn{5}{l}{\textbf{\textit{Z-S w/ Strategy 2}:}} \\
Qwen-2.5-VL (arXiv'25)            & 12.91 & 6.08  & 7.07  & 4.58 \\
\multirow{2}{*}{\textbf{Video RSVG-ZeroOV (Ours)}} 
& \underline{28.35} & 11.29 & 12.98 & 7.60 \\
& {\scriptsize\textcolor{darkgreen}{$\uparrow$15.44}} 
& {\scriptsize\textcolor{darkgreen}{$\uparrow$5.21}} 
& {\scriptsize\textcolor{darkgreen}{$\uparrow$5.91}} 
& {\scriptsize\textcolor{darkgreen}{$\uparrow$3.02}} \\
\specialrule{.1em}{.3em}{.3em}
\end{tabular}}
\label{tab:VidSTG}
\end{table}

\subsubsection{Performance on General Video Grounding}

\noindent \textbf{Comparison Results on HC-STVG.}
To evaluate the generalization ability of Video RSVG-ZeroOV beyond remote sensing videos, we further conduct experiments on the general-domain HC-STVGv1 and HC-STVGv2 benchmarks. As shown in Table~\ref{tab:HC_STVG}, our method consistently outperforms existing zero-shot baselines on both datasets, demonstrating that the proposed framework is not limited to aerial videos but can generalize to broader spatio-temporal grounding scenarios.
With Strategy 1, Video RSVG-ZeroOV achieves the best zero-shot performance on HC-STVGv1 in terms of m\_vIoU and vIoU@0.3, improving the strongest existing zero-shot baseline by 3.56 and 5.01 points, respectively. On HC-STVGv2, Strategy 1 obtains the best zero-shot results across all reported metrics, improving Qwen-2.5-VL by 10.21 points in m\_vIoU, 12.82 points in vIoU@0.3, and 7.29 points in vIoU@0.5.
Compared with weakly-supervised methods, Video RSVG-ZeroOV also achieves competitive or superior results. On HC-STVGv1, Strategy 1 surpasses all weakly-supervised baselines in m\_vIoU and vIoU@0.3, and achieves the second-best result on vIoU@0.5. On HC-STVGv2, it obtains the best results in m\_vIoU and vIoU@0.3 among weakly-supervised and zero-shot methods, while remaining close to the strongest weakly-supervised result on vIoU@0.5. This is notable because weakly-supervised methods still rely on task-related training data, whereas our method performs zero-shot inference using frozen foundation models.

\noindent \textbf{Comparison Results on VidSTG.}
Table~\ref{tab:VidSTG} reports the comparison results on the VidSTG test set.
Video RSVG-ZeroOV with Strategy 1 achieves the best zero-shot performance across all reported metrics, including temporal localization, video-level overlap, and high-threshold vIoU.
Compared with the strongest existing zero-shot baseline, our method improves m\_vIoU by 1.50 points, vIoU@0.3 by 0.65 points, and vIoU@0.5 by 3.35 points.
For temporal localization, Video RSVG-ZeroOV substantially outperforms Qwen-2.5-VL, improving m\_tIoU by 17.25 points over the stronger Qwen-2.5-VL variant.
These results demonstrate that query-aware key-frame grounding and SAM3-based temporal propagation can generalize to diverse general-domain videos.

Video RSVG-ZeroOV also surpasses weakly-supervised methods on the main vIoU metrics.
Specifically, Strategy 1 improves over the strongest weakly-supervised results by 1.71 points in m\_vIoU, 1.02 points in vIoU@0.3, and 2.16 points in vIoU@0.5.
Although fully-supervised methods remain stronger on VidSTG due to task-specific video annotations, the strong zero-shot performance of Video RSVG-ZeroOV shows that training-free video grounding can achieve competitive spatial-temporal localization when reliable key-frame grounding and temporal propagation are combined.

\subsection{Ablation Study and Analysis}
\label{sec4c}

\subsubsection{Component Analysis of RSVG-ZeroOV.}
As shown in Table~\ref{tab:ablation1}, removing VLM (replaced by a cross-attention map in DM) leads to significant drops of \textbf{14.10\%} and \textbf{6.04\%} in mIoU on both tasks.
Excluding DM results in failure on RSRES, as the model loses structural priors necessary for segment object regions.
Exp.(\textit{w/o} Evolve) highlight the benefit of attention evolve module, increasing the Pr@0.5 and mIoU of \textbf{7.52\%} and \textbf{6.27\%} on RSREC.
The improved performance of O-F-E over O-E-F indicates that embedding the DM's self-attention into the VLM's cross-attention better preserves structural and shape information of objects, which facilitates more accurate region aggregation via DFS and suppresses noisy activations.

\subsubsection{Effect of Self-Attention Resolution.}
As shown in Table~\ref{tab:ablation2}, we investigate the impact of varying resolutions of self-attention maps.
We find that using multiple resolutions [32, 64] leads to the best performance, suggesting that multi-scale self-attention maps not only preserve high-resolution details but also effectively capture contextual semantics.

\begin{table}[!t]
\centering
\caption{Component ablation of RSVG-ZeroOV on RRSIS-D.}
\resizebox{\linewidth}{!}{
\begin{tabular}{l|cc|cc}
    \specialrule{.1em}{.3em}{.3em} 
    \multicolumn{1}{c|}{}       & \multicolumn{2}{c|}{RSREC} & \multicolumn{2}{c}{RSRES}\\
    \cmidrule(r){2-5}       
    \multicolumn{1}{l|}{Method} & Pr@0.5 & mIoU & Pr@0.5 & mIoU\\
    \specialrule{.1em}{.3em}{.3em} 
    RSVG-ZeroOV (Ours)       & \textbf{30.15} & \textbf{32.92} & \textbf{12.84} & \textbf{21.85}   \\
    \textit{w/o} VLM         & 16.22 & 18.82 & 11.43 & 15.81 \\
    \textit{w/o} DM          & 21.49 & 26.26 &  1.18 &  6.15 \\
    \textit{w/o} Evolve      & 22.63 & 26.65 & 10.26 & 20.56 \\
    \midrule
    \textit{O-F-E} Stage   & \textbf{30.15} & \textbf{32.92} & \textbf{12.84} & \textbf{21.85}\\
    \textit{O-E-F} Stage   & 27.34 & 29.51 & 7.18 & 15.89\\
    \specialrule{.1em}{.3em}{.3em} 
\end{tabular}}
\label{tab:ablation1}
\end{table}

\begin{table}[!t]
\centering
\caption{Effect of self-attention map resolution on RRSIS-D.}
\setlength{\tabcolsep}{2.2mm}
{\fontsize{9pt}{9.5pt}\selectfont
\begin{tabular}{c|cc|cc}
    \specialrule{.1em}{.3em}{.3em} 
    \multicolumn{1}{c|}{}                        & \multicolumn{2}{c|}{RSREC} & \multicolumn{2}{c}{RSRES}\\
    \cmidrule(r){2-5}       
    \multicolumn{1}{l|}{Resolution}  & Pr@0.5 & mIoU & Pr@0.5 & mIoU\\
    \specialrule{.1em}{.3em}{.3em}
    16              & 26.16 & 28.39 & 12.01 & 19.98 \\
    32              & 29.84 & 31.97 & 12.53 & 21.11   \\
    64              & 27.70 & 30.76 & 10.55 & 20.13 \\
    16, 32          & 28.23 & 29.86 & 12.70 & 20.41 \\
    32, 64          & \textbf{30.15} & \textbf{32.92} & \textbf{12.84} & \textbf{21.85}   \\
    16, 32, 64      & 27.86 & 30.51 & 11.98 & 20.36 \\
    \specialrule{.1em}{.3em}{.3em} 
\end{tabular}}
\label{tab:ablation2}
\end{table}

\begin{figure}[!t]
    \centering
    \includegraphics[width=\linewidth]{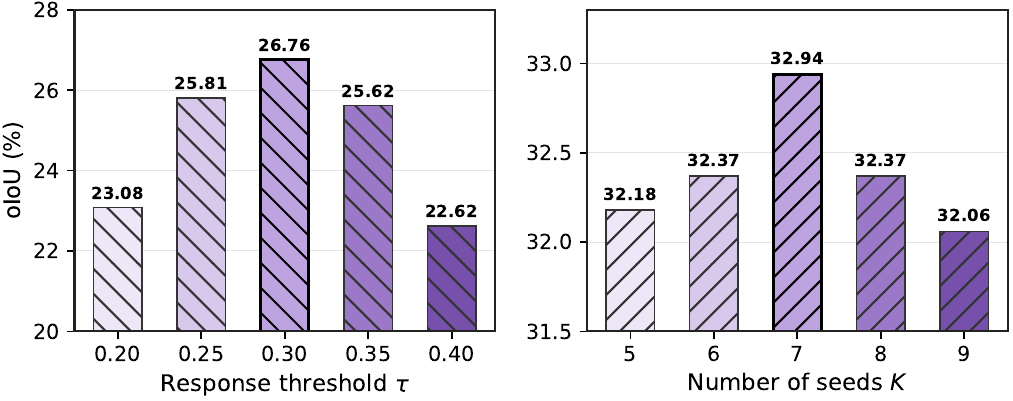}
    \caption{Effect of $\tau$ and $K$ in the Evolve stage on RRSIS-D.}
    \label{fig:ablation_tau_k}
\end{figure}

\subsubsection{Hyperparameter Analysis of the Evolve Stage.} 
We conducted ablation studies on two key hyperparameters in the Evolve stage: the number of top-K seeds and the response threshold $\tau$. First, with $K$ fixed at 1, we evaluated the impact of different $\tau$ values. As shown in Figure \ref{fig:ablation_tau_k}, the oIoU for the RSRES task reaches its peak value of 26.76\% when $\tau$ is set to 0.3. Then, keeping $\tau$ fixed at 0.3, we explored the effect of varying $K$. As shown, the oIoU reaches its maximum of 32.94\% when $K$ is set to 7. These results confirm that well-chosen seed quantities and response thresholds can significantly enhance the performance of the Evolve stage.

\subsubsection{Effect of Mask Binarization Threshold.}
As shown in Table \ref{tab:threshold_refinement}, as the threshold increases from 0.2 to 0.5, the performance on both RSREC and RSRES tasks exhibits a ”rise-then-fall” pattern. This is primarily because a low threshold introduces excessive background noise, leading to overgeneralization in the predicted regions. In contrast, a high threshold may exclude partial boundary areas, thereby reducing recall and spatial coverage. To balance performance between the RSREC (mIoU 32.92\%) and RSRES (mIoU 21.85\%) tasks, we select $\alpha$= 0.4 as the default threshold, as it achieves balanced performance across both tasks, ensuring segmentation consistency while improving model robustness.

\subsubsection{Analysis of Optional Mask Refinement.} 
We evaluate several representative refinement strategies, including the traditional DenseCRF post-processing method and different prompt mechanisms based on SAM. Table \ref{tab:threshold_refinement} shows that the SAM (Box) refinement strategy achieves the best performance on both tasks, with mIoU scores of 34.49\% for RSREC and 28.35\% for RSRES, significantly outperforming other methods.

\begin{table}[!t]
\centering
\caption{Ablation study on mask binarization thresholds and refinement strategies on RRSIS-D.}
\resizebox{\linewidth}{!}{
\begin{tabular}{l|cc|cc}
    \specialrule{.1em}{.3em}{.3em}
    \multicolumn{1}{c|}{}       & \multicolumn{2}{c|}{RSREC} & \multicolumn{2}{c}{RSRES}\\
    \cmidrule(r){2-5}       
    \multicolumn{1}{l|}{Method} & Pr@0.5 & mIoU & Pr@0.5 & mIoU\\
    \specialrule{.1em}{.3em}{.3em}
    RSVG-ZeroOV (Ours) & 30.15 & 32.92 & \textbf{12.84} & 21.85 \\
    \specialrule{.08em}{.25em}{.25em}
    \multicolumn{5}{l}{Mask Threshold $\alpha$} \\
    0.2 & 26.10 & 29.24 & 10.69 & 21.35 \\
    0.3 & 28.23 & 29.86 & 12.70 & 22.41 \\
    0.4 & 30.15 & 32.92 & \textbf{12.84} & 21.85 \\
    0.5 & \textbf{32.22} & \textbf{33.79} & 10.92 & 19.46 \\
    0.6 & 29.59 & 31.28 & 7.70 & 15.87 \\
    \specialrule{.08em}{.25em}{.25em}
    \multicolumn{5}{l}{Refinement Methods} \\
    + DenseCRF & 30.44 & 34.17 & 15.36 & 23.00 \\
    + SAM (SMR) & 28.95 & 31.37 & 26.62 & 27.68 \\
    + SAM (Point) & 3.58 & 8.86 & 3.58 & 6.23 \\
    + SAM (Box) & \textbf{31.39} & \textbf{34.49} & \textbf{27.39} & \textbf{28.35} \\
    + SAM (Box+Point) & 28.26 & 32.04 & 23.26 & 26.78 \\
    \specialrule{.1em}{.3em}{.3em}
\end{tabular}}
\label{tab:threshold_refinement}
\end{table}

\begin{figure}[!t]
    \centering
    \includegraphics[width=\linewidth]{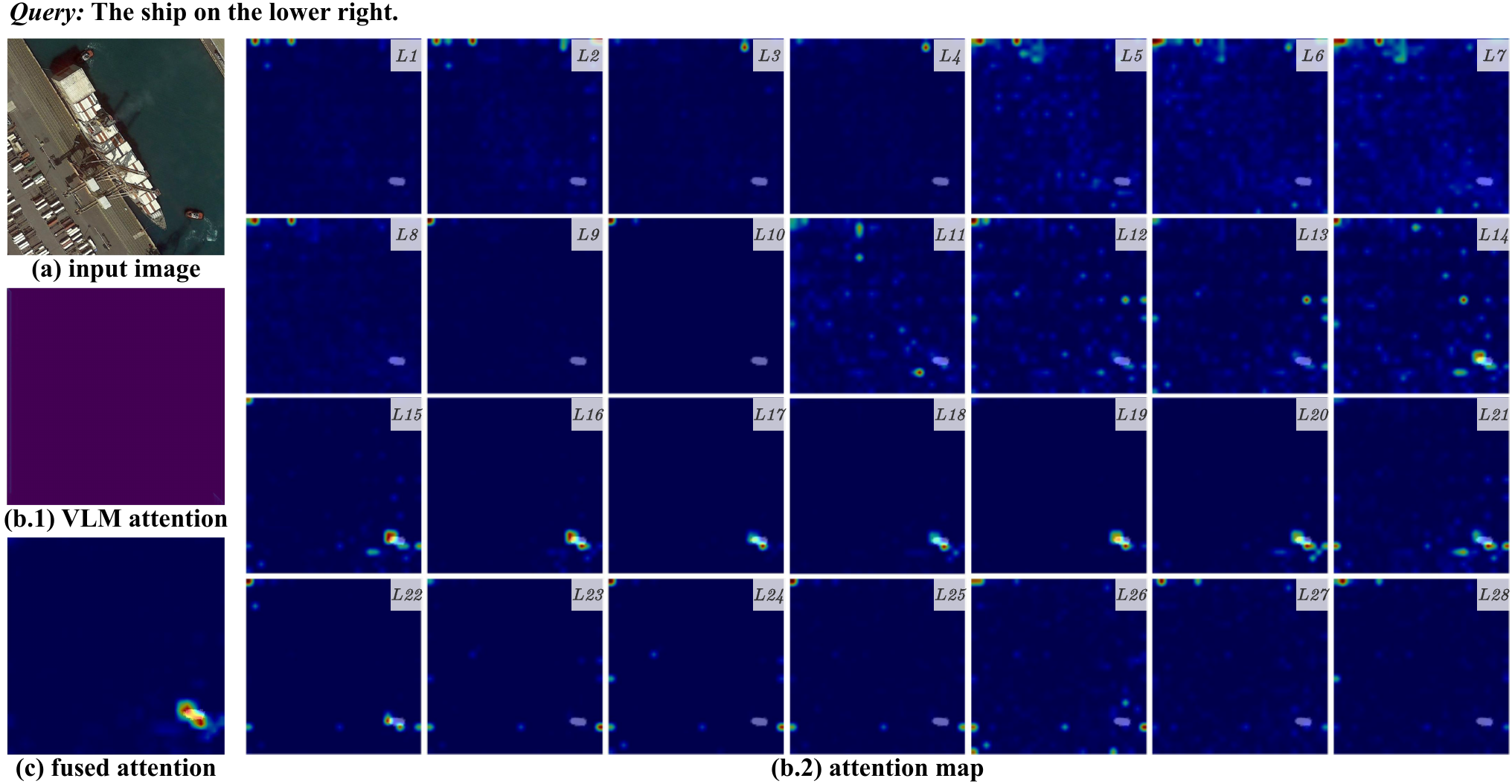}
    \caption{Analysis of VLM attention selection. 
    (a) Input image and query. 
    (b.1) Overall VLM attention response. 
    (b.2) Layer-wise image-text-related attention maps extracted from the VLM, showing distinct localization behaviors across layers. 
    (c) Fused attention map obtained by weighted aggregation of selected middle-to-late layers in (b.2).}
    \label{fig:vlm_attention_selection}
\end{figure}

\subsubsection{Analysis of VLM Attention Selection.}
We further analyze how to extract reliable localization cues from the frozen VLM in the Overview stage. 
For decoder-only VLMs, the attention matrix contains interactions among visual tokens, textual tokens, and generated output tokens. 
As shown in Fig.~\ref{fig:vlm_attention_selection} \textcolor{red}{(b.1)}, directly visualizing the overall attention response does not explicitly reveal which visual regions are responsible for grounding the referring expression. 
Therefore, we extract the image-text-related attention maps associated with visual tokens during language generation, and visualize their layer-wise responses in Fig.~\ref{fig:vlm_attention_selection} \textcolor{red}{(b.2)}.
As shown in Fig.~\ref{fig:vlm_attention_selection} \textcolor{red}{(b.2)}, different layers exhibit distinct localization behaviors. 
Early layers tend to produce broad or scattered responses, indicating limited language-conditioned selectivity. 
Middle layers begin to highlight object-level regions, while later layers provide stronger semantic responses to the referred target but may focus on only part of the object. 
This layer-dependent behavior is consistent with recent findings that reliable grounding cues in large vision-language models are concentrated in only a subset of layers~\cite{kang2025your}.
Based on this observation, we obtain the final VLM attention prior by fusing selected middle-to-late layer maps from Fig.~\ref{fig:vlm_attention_selection} \textcolor{red}{(b.2)}. 
Specifically, we aggregate the visual-token attention maps from layers 16, 17, 18, and 19 with weights of 0.1, 0.1, 0.3, and 0.5, respectively. 
The resulting fused attention map, shown in Fig.~\ref{fig:vlm_attention_selection} \textcolor{red}{(c)}, better highlights the referred object and provides a stable coarse localization prior for the subsequent \textit{Focus} and \textit{Evolve} stages.

\begin{figure}[!t]
    \centering
    \includegraphics[width=\linewidth]{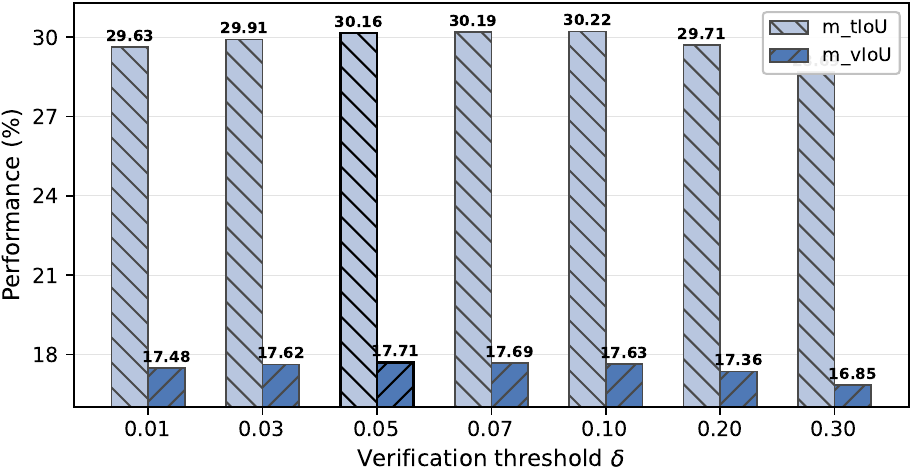}
    \caption{Ablation study on the verification threshold $\delta$ on VidSTG.}
    \label{fig:ablation_delta}
\end{figure}

\begin{table}[t]
\centering
\caption{Comparison of different SAM variants for video-level temporal propagation on UAV-SVG.}
\label{tab:sam_ablation_image_video}
\resizebox{\linewidth}{!}{
\begin{tabular}{lcccc}
\toprule
\multirow{2}{*}{Metric} 
& \multicolumn{2}{c}{Strategy 1} 
& \multicolumn{2}{c}{Strategy 2} \\
\cmidrule(lr){2-3} \cmidrule(lr){4-5}
& SAM2 & SAM3 & SAM2 & SAM3 \\
\midrule
m\_vIoU  & 24.93 & 27.19 {\scriptsize\textcolor{darkgreen}{$\uparrow$2.26}} & 19.29 & 21.96 {\scriptsize\textcolor{darkgreen}{$\uparrow$2.26}} \\
vIoU@0.3 & 33.96 & 35.62 {\scriptsize\textcolor{darkgreen}{$\uparrow$1.66}} & 25.93 & 27.91 {\scriptsize\textcolor{darkgreen}{$\uparrow$1.98}} \\
vIoU@0.5 & 26.85 & 29.28 {\scriptsize\textcolor{darkgreen}{$\uparrow$2.43}} & 17.65 & 20.80 {\scriptsize\textcolor{darkgreen}{$\uparrow$3.15}} \\
m\_fAcc  & 29.18 & 33.77 {\scriptsize\textcolor{darkgreen}{$\uparrow$4.59}} & 21.15 & 24.57 {\scriptsize\textcolor{darkgreen}{$\uparrow$3.42}}\\
fAcc@0.3 & 35.34 & 37.87 {\scriptsize\textcolor{darkgreen}{$\uparrow$2.53}} & 25.61 & 28.78 {\scriptsize\textcolor{darkgreen}{$\uparrow$3.17}}\\
fAcc@0.5 & 30.52 & 34.61 {\scriptsize\textcolor{darkgreen}{$\uparrow$4.09}} & 22.08 & 24.84 {\scriptsize\textcolor{darkgreen}{$\uparrow$2.76}}\\
\bottomrule
\end{tabular}}
\end{table}

\subsubsection{Effect of VLM-Guided Temporal Verification.}
We analyze the effect of the verification threshold $\delta$ in the VLM-guided temporal verification module on VidSTG. 
As shown in Figure~\ref{fig:ablation_delta}, both m\_tIoU and m\_vIoU remain stable within a small threshold range around $\delta=0.05$, indicating that the proposed verification mechanism is not sensitive to minor threshold variations. 
When $\delta$ becomes too large, the verification process tends to reject valid frames, leading to degraded temporal grounding performance. 
Therefore, we set $\delta=0.05$ as the default threshold in all experiments.

\subsubsection{Analysis of SAM Variants for Video Propagation.}
As shown in Table~\ref{tab:sam_ablation_image_video}, replacing SAM2 with SAM3 consistently improves all metrics under both strategies, confirming the stronger temporal propagation ability of SAM3. 
More importantly, even when using SAM2, Video RSVG-ZeroOV with Strategy 1 still achieves 24.93 m\_vIoU and 26.85 vIoU@0.5, outperforming most fully-supervised methods in Table~\ref{tab:UAV-SAVG}. 
This indicates that the effectiveness of our framework does not solely come from SAM3, but also from the query-aware key-frame grounding and tube selection design.

\section{Conclusion}
\label{sec:conclusion}
In this paper, we presented RSVG-ZeroOV, a training-free framework for zero-shot open-vocabulary remote sensing visual grounding in both images and videos. 
The design of RSVG-ZeroOV is motivated by the complementary strengths of frozen generic foundation models: VLMs provide strong semantic alignment between free-form language expressions and visual regions, while DMs encode object-centric structural priors that are useful for recovering complete object extents.
Based on this observation, we introduced an \textit{Overview--Focus--Evolve} strategy to progressively refine attention maps for accurate localization of referred objects.
To extend RSVG-ZeroOV from images to videos, we further developed Video RSVG-ZeroOV, which combines a query-relevant key-frame selector with a SAM3-based temporal propagator for training-free spatio-temporal grounding. 
Extensive experiments on six image and video grounding benchmarks demonstrate that the proposed framework consistently outperforms existing zero-shot baselines and achieves competitive or superior performance compared with weakly- or fully-supervised methods in several settings. 
These results demonstrate the feasibility and potential of using frozen generic foundation models for open-world remote sensing visual grounding, and we hope this work will inspire further research on more capable remote sensing vision-language models and more general open-world geospatial perception systems.

\bibliographystyle{IEEEtran}
\bibliography{ref.bib}

\end{document}